\documentclass[11pt]{article}

\usepackage{amsmath}
\usepackage{amsfonts}
\usepackage{amssymb}
\usepackage{booktabs}
\usepackage{makecell}
\usepackage{multirow} 
\usepackage{subcaption}
\usepackage{bm}
\usepackage{array}
\usepackage{colortbl}
\usepackage{xcolor}
\usepackage{booktabs}
\usepackage[hyphens]{url}

\usepackage[final]{acl}

\usepackage{times}
\usepackage{latexsym}

\usepackage[T1]{fontenc}

\usepackage[utf8]{inputenc}

\usepackage{microtype}

\usepackage{inconsolata}

\usepackage{graphicx}

\definecolor{deductive}{RGB}{61,133,198}
\definecolor{inductive}{RGB}{106,168,79}
\definecolor{abductive}{RGB}{255,0,0}
\newcommand{\hlD}[1]{\colorbox{deductive!30}{#1}}
\newcommand{\hlI}[1]{\colorbox{inductive!30}{#1}}
\newcommand{\hlA}[1]{\colorbox{abductive!30}{#1}}

\title{Knowledge Vector of Logical Reasoning in Large Language Models}

\author{Zixuan Wang and Yuanyuan Lei\\
        Department of Computer \& Information Science and Engineering\\
        University of Florida, Gainesville, FL, United States\\
        \texttt{\{zwang10, yuanyuan.lei\}@ufl.edu}}

\begin{document}
\maketitle
\begin{abstract}

Logical reasoning serve as a central capability in LLMs and includes three main forms: deductive, inductive, and abductive reasoning. In this work, we study the knowledge representations of these reasoning types in LLMs and analyze the correlations among them. Our analysis shows that each form of logical reasoning can be captured as a reasoning-specific knowledge vector in a linear representation space, yet these vectors are largely independent of each other. Motivated by cognitive science theory that these subforms of logical reasoning interact closely in the human brain, as well as our observation that the reasoning process for one type can benefit from the reasoning chain produced by another, we further propose to refine the knowledge representations of each reasoning type in LLMs to encourage complementarity between them. To this end, we design a complementary subspace-constrained refinement framework, which introduces a complementary loss that enables each reasoning vector to leverage auxiliary knowledge from the others, and a subspace constraint loss that prevents erasure of their unique characteristics. Through steering experiments along reasoning vectors, we find that refined vectors incorporating complementary knowledge yield consistent performance gains. We also conduct a mechanism-interpretability analysis of each reasoning vector, revealing insights into the shared and specific features of different reasoning in LLMs\footnote{The code link is: \url{https://github.com/lei-nlp-lab/knowledge_vector_acl_2026}}.

\end{abstract}

\section{Introduction}

\begin{figure}[htbp]
    \centering
    \includegraphics[width=\linewidth]{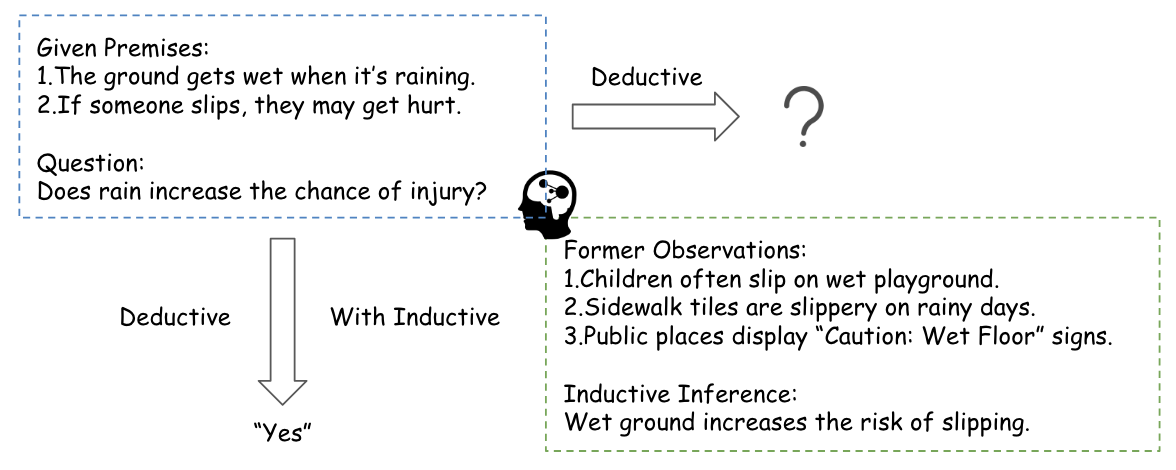}
    \caption{An example showing how multiple logical reasoning stratepicturesgies cooperate in human thinking. Rather than operating in isolation, inductive reasoning contributes newly formed premises derived from former observations, which are incorporated into a deductive reasoning process to reach a final conclusion.}
    \label{introduction_example}
\end{figure}


Despite the impressive capabilities of Large Language Models (LLMs) across diverse tasks \citep{openai2024gpt4technicalreport, guan2023leveraging, yao2022react}, understanding their reasoning abilities, particularly logical reasoning \citep{pan2023logic, lam2024closer}, which is the foundation of problem solving, remains poorly understood and requires deeper investigation. Logical reasoning is commonly organized into three principal categories in standard logic theory \cite{peirce1934collected}: abductive, deductive, and inductive reasoning.
In practice, it has become a new understanding perspective being applied to different areas, such as procedural planning \cite{kang2025exploring}, coding \cite{vashishtha2025executable}, and mathematics \cite{abdaljalil2025theorem}, where the model must draw reliable conclusions or hypotheses from structured information rather than relying solely on surface correlations \cite{bedi2025fidelity}. Therefore, understanding how LLMs perform logical reasoning is essential for building models that are trustworthy, generalizable, and aligned with human expectations of rational inference.

At the same time, knowledge vectors \citep{rimsky2024steering, turner2023steering} have been proven to be meaningful under the assumption of linear representation in a model's activation space \citep{jorgensen2023improving, zou2023representation}. Prior work suggests that certain knowledge-related concepts can be encoded linearly, such as truthfulness \cite{wang2025adaptive}, instruction-following \cite{stolfo2024improving} etc. However, no prior works study whether reasoning abilities, particularly different forms of logical reasoning, can be linearly represented in the model's activation space. The most closely related line of work is \citet{venhoff2025understanding}, which investigates knowledge vectors for specific reasoning patterns such as backtracking, but does not address reasoning capabilities in a more general sense.

To investigate whether different types of logical reasoning can be linearly represented within LLMs, we extract a knowledge vector for each reasoning type and examine whether the knowledge-enhancement effect emerges as well \cite{nanda2023othello}. By using these vectors to steer the model's behavior, we observe improved reasoning performance corresponding to each reasoning vector.


However, further geometric analysis reveals low pariwise cosine similarities among these naively extracted reasoning vectors, which means different types of reasoning knowledge are represented as largely distinct representation in the model's activation space. This observation contrasts with findings in human cognitive science \citep{holyoak2013oxford, heit2010relations}, which suggest that different forms of logical reasoning share common components and interact in a complementary manner in human thinking process. Take Figure~\ref{introduction_example} as an illustration, inductive reasoning can support deductive reasoning by contributing newly formed premises derived from prior observations, thereby facilitating the derivation of a final conclusion.

Inspired by the above observations, we propose a complementary refinement method for enhancing reasoning knowledge, by encouraging the model to mimic human reasoning paradigm. Sparse Autoencoders (SAEs) have been shown to reliably interpret semantic features within LLMs by disentangling the complex, superimposed features into more interpreterable components \cite{shu2025survey}. Leveraging this property, we introduce a complementary subspace-constrained refinement framework, consisting of (i) a complementary loss that enables each reasoning vector to incorporate complementary reasoning knowledge from the other two, thereby refining the resulting knowledge vectors, and (ii) a subspace constraint loss that restricts each reasoning vector to its corresponding SAE-induced subspace, preventing the erasure of its unique characteristics. Take Llama-3.1-8B-it as an example, by using the complementary vectors, we observed the metric boost $55.22-56.46$, $27.13-27.55$, $39.10-42.07$ for deductive, inductive and abductive reasoning respectively.

To better understand the model's internal reasoning mechanisms, we further conduct a fine-grained mechanism interpretation analysis. Firstly, we examine SAE-extracted feature sets across three types of logical reasoning. Our analysis reveals the emergence of complementary features acquired from other reasoning types through the refinement process. We also observe that deductive and inductive reasoning becomes more closely aligned while abductive becomes more specialized, which is consistent with the fact that deductive and inductive reasoning share a similar evidence-based reasoning style. Secondly, we employ attention patching \cite{heimersheim2024use} to analyze the model's internal circuit before and after refinement. Our finding is that core activations are largely preserved, activation patterns become more concentrated, and new activations emerge following the refinement process. Thirdly, our qualitative analysis identifies key text spans associated with each type of logical reasoning, illustrating how the model allocates attention across different reasoning processes \cite{han2024word}.

\section{Preliminary Exploration}

\begin{figure*}[t]
    \centering
    \includegraphics[width=\textwidth]{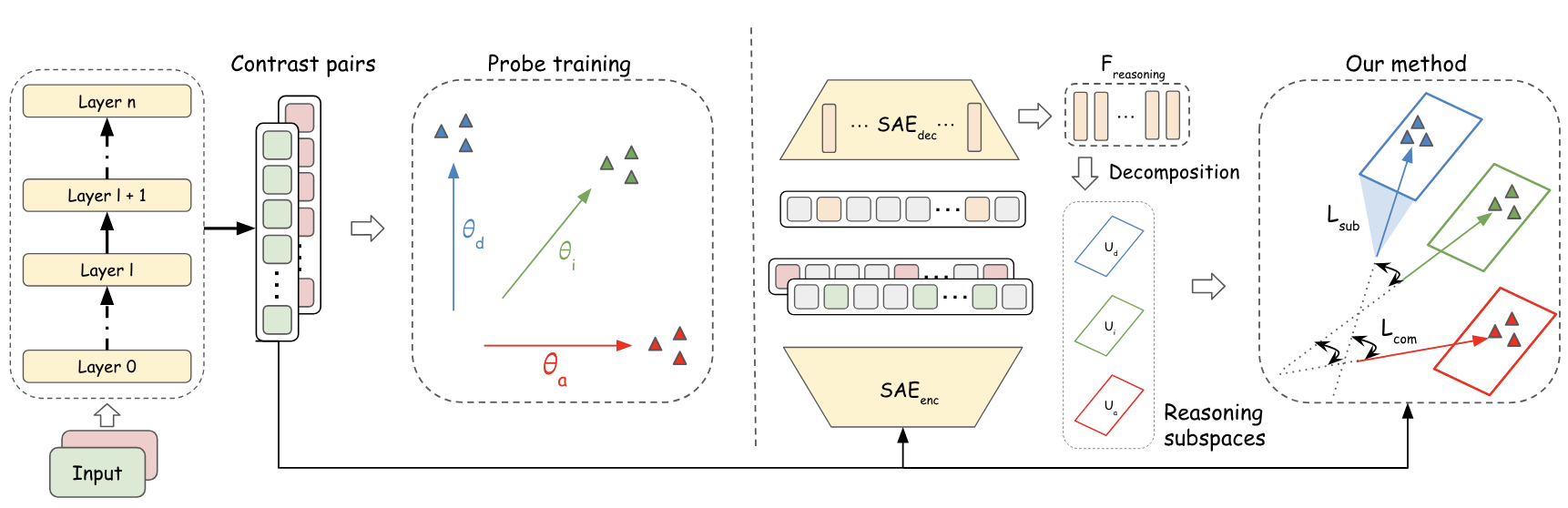}
    \caption{Comparison between the naive and complementary extraction. The left panel illustrates the naive approach, where contrast pairs are collected and used to train linear probes that serve as reasoning vectors. $\theta_{d}$, $\theta_{i}$ and $\theta_{a}$ are reasoning vectors for deductive, inductive and abductive reasoning respectively. The right panel presents our complementary method, where we first extract highly related reasoning features $F_{reasoning}$ from SAE and construct a subspace for each type of logical reasoning via QR decomposition. A complementary knowledge loss $L_{com}$ is designed, together with a subspace constraint loss $L_{sub}$ based on SAE-induced subspace.}
    \label{fig:fig3}
\end{figure*}

\begin{figure}[t]
    \centering
    \includegraphics[width=0.7\linewidth]{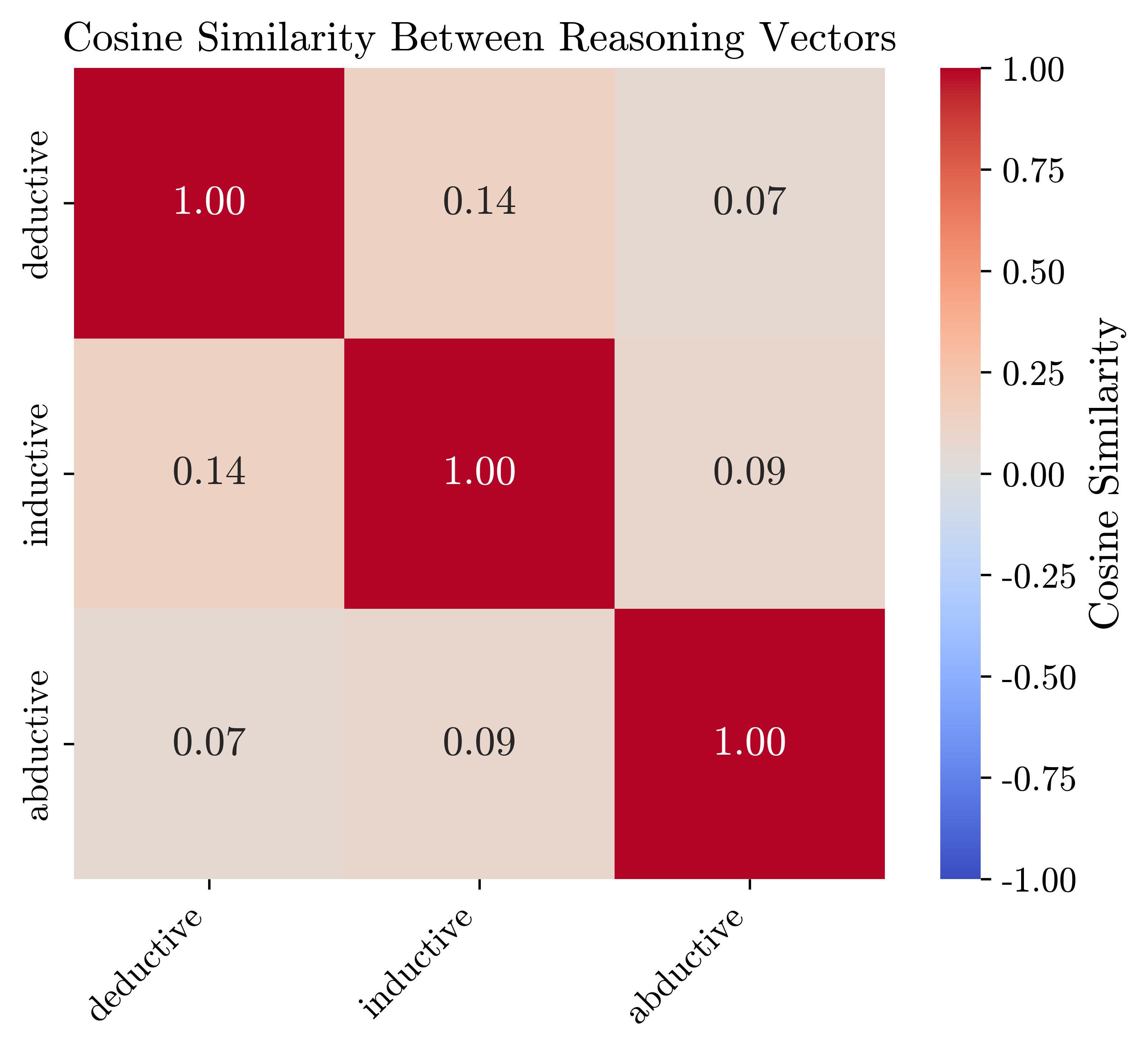}
    \caption{Cosine similarity heatmaps between knowledge vectors for different logical reasoning. The values range from $-1$ to $1$, where negative values denote opposing directions and positive values indicate alignment. The generally near-zero scores indicate that the learned vectors encode distinct reasoning patterns in model's activation space.}
    \label{fig:fig1}
\end{figure}






\begin{table*}[t]
\centering
\small
\resizebox{\textwidth}{!}{
\begin{tabular}{c ccc ccc ccc}
\toprule
\textbf{Method}  
    & \multicolumn{3}{c}{\textbf{Llama-3.1-8B-it}} 
    & \multicolumn{3}{c}{\textbf{Gemma-2-9B-it}} 
    & \multicolumn{3}{c}{\textbf{GPT-OSS-20B}} \\
\cmidrule(lr){2-4} \cmidrule(lr){5-7} \cmidrule(lr){8-10}
    & Deductive & Inductive & Abductive 
    & Deductive & Inductive & Abductive 
    & Deductive & Inductive & Abductive \\
\midrule

\multicolumn{10}{c}{\textbf{Greedy}} \\
\midrule

Unsteered
 & $48.95$ & $26.36$ & $32.27$
 & $56.86$ & $24.52$ & $54.67$ 
 & $56.12$ & $16.68$ & $45.09$ \\

Mono Steering
 & $55.22$ & $27.13$ & $39.19$
 & $57.33$ & $26.49$ & $55.74$ 
 & $57.63$ & $17.24$ & $47.19$ \\

Complementary Steering
 & $\bm{56.46}$ & $\bm{27.55}$ & $\bm{40.95}$
 & $\bm{59.05}$ & $\bm{27.03}$ & $\bm{58.20}$ 
 & $\bm{58.43}$ & $\bm{18.52}$ & $\bm{50.50}$ \\

\midrule\midrule

\multicolumn{10}{c}{\textbf{Sampling@5}} \\
\midrule

Unsteered
 & $47.45$ & $24.25$ & $39.01$
 & $49.74$ & $23.85$ & $54.92$ 
 & $51.12$ & $22.91$ & $41.69$ \\

Mono Steering
 & $49.90$ & $24.48$ & $39.10$
 & $50.69$ & $24.51$ & $55.98$ 
 & $52.08$ & $23.16$ & $42.13$ \\

Complementary Steering
 & $\bm{51.36}$ & $\bm{24.78}$ & $\bm{42.07}$
 & $\bm{54.64}$ & $\bm{25.56}$ & $\bm{57.53}$ 
 & $\bm{53.67}$ & $\bm{23.92}$ & $\bm{46.89}$ \\

\bottomrule
\end{tabular}
}
\caption{Performance comparison under Greedy and Sampling@5 decoding settings. 
Greedy is the default decoding strategy, while Sampling@5 denotes averaging results over five sampled generations. Complementary steering consistently achieves the strongest performance.}
\label{table:performance}
\end{table*}

\begin{figure*}
\centering

\begin{minipage}{0.32\linewidth}
    \centering
    \includegraphics[width=\linewidth]{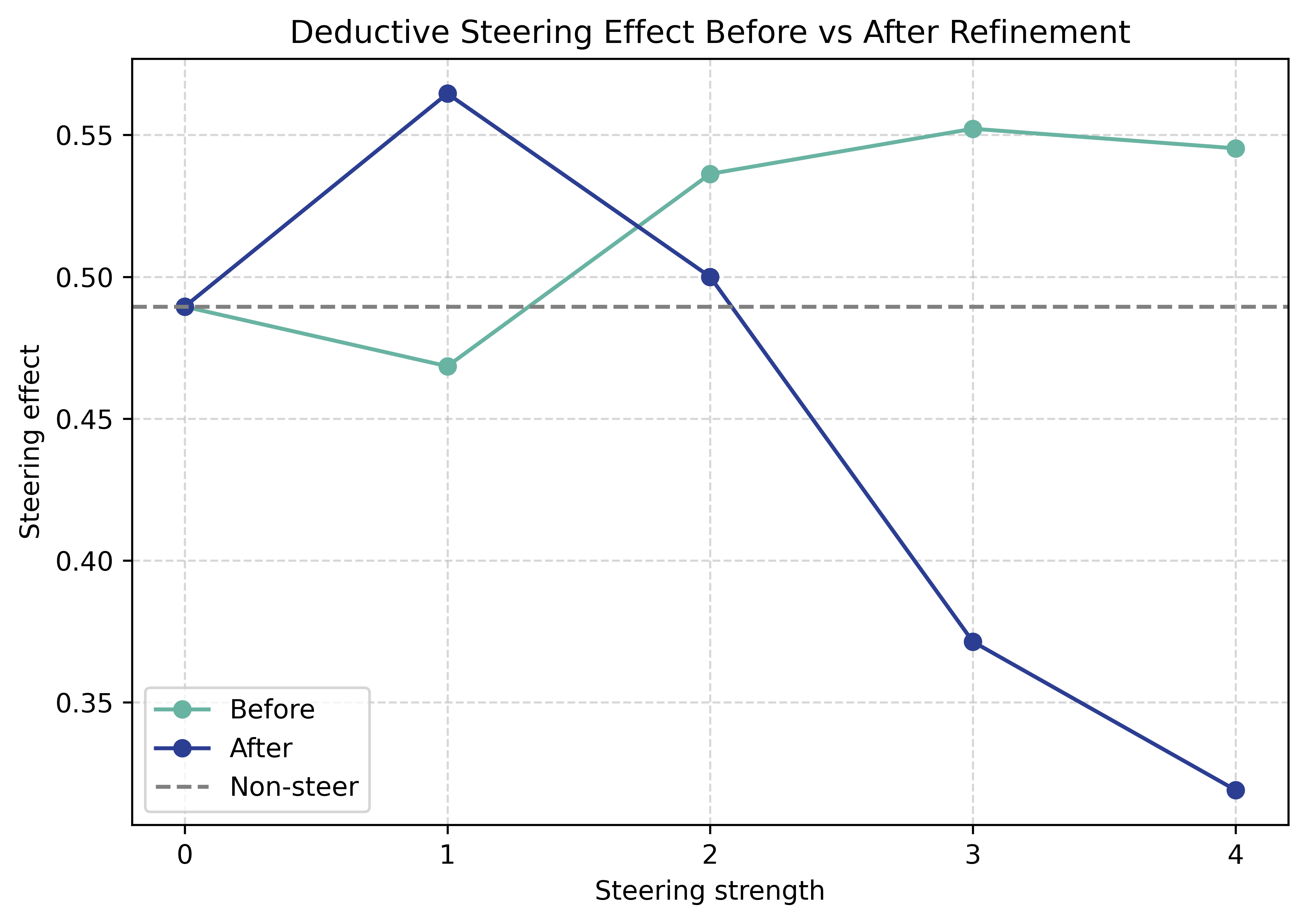}
    \caption*{(a) Deductive}
\end{minipage}
\hfill
\begin{minipage}{0.32\linewidth}
    \centering
    \includegraphics[width=\linewidth]{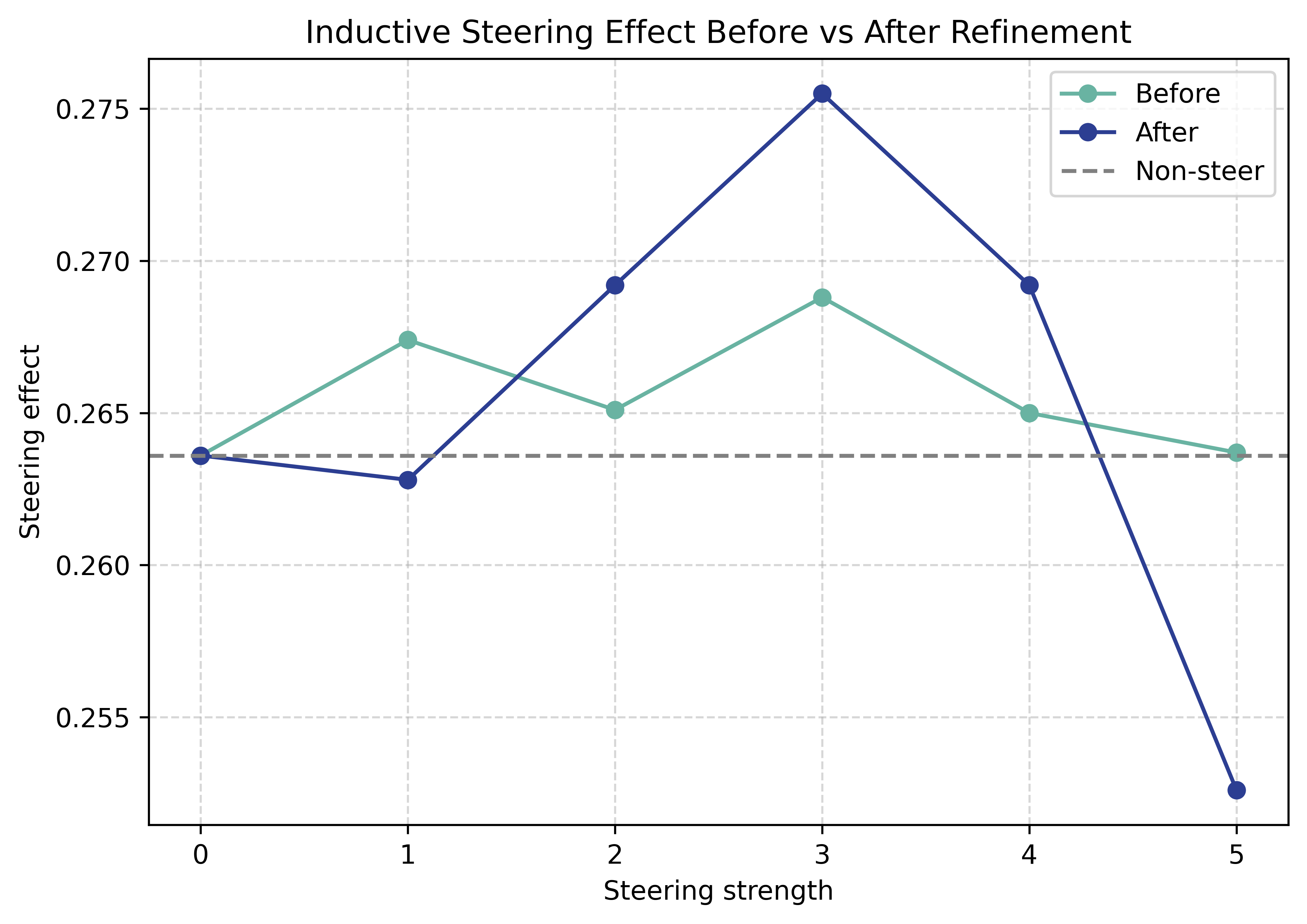}
    \caption*{(b) Inductive}
\end{minipage}
\hfill
\begin{minipage}{0.32\linewidth}
    \centering
    \includegraphics[width=\linewidth]{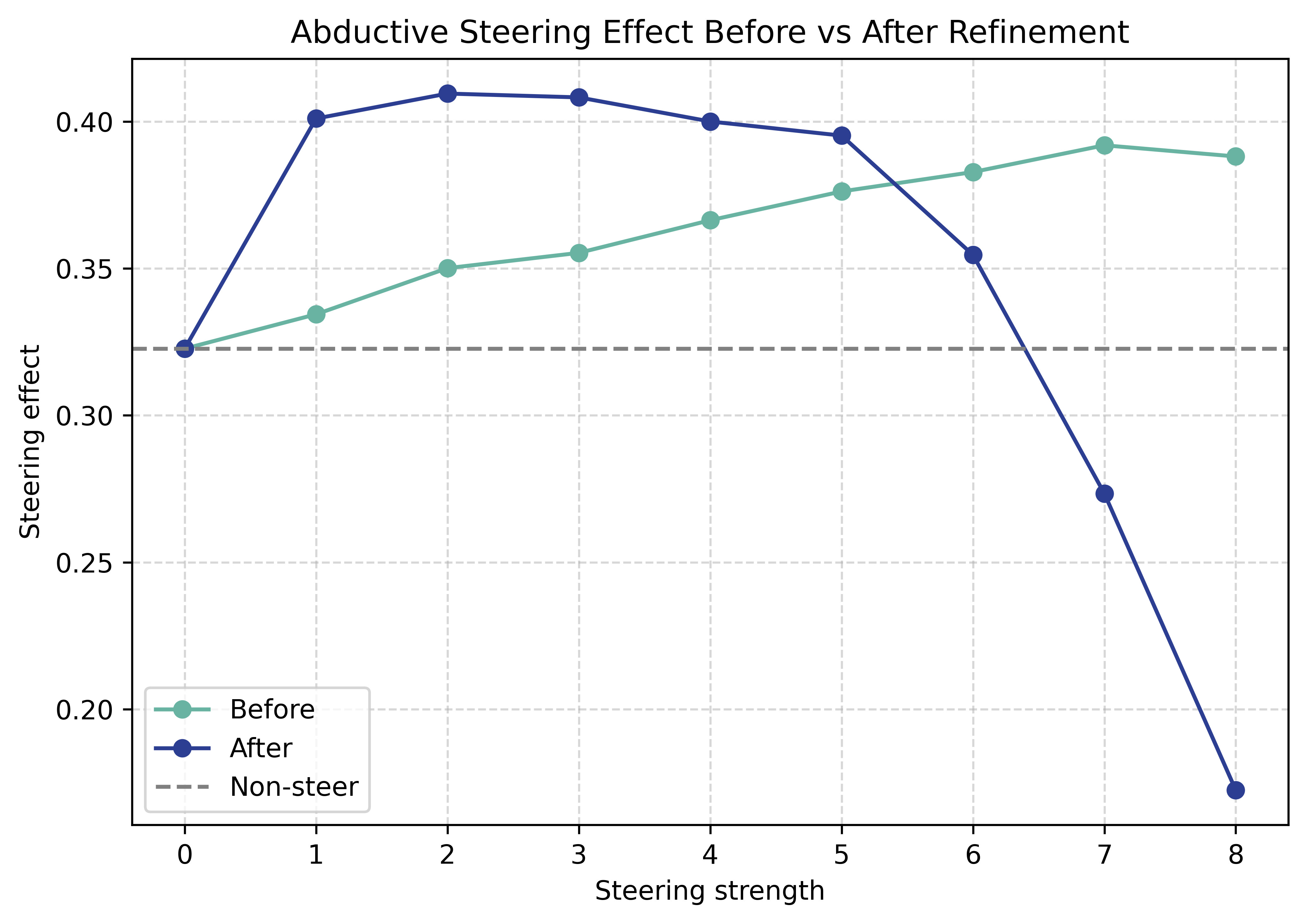}
    \caption*{(c) Abductive}
\end{minipage}

\caption{The effect of different steering strength before and after complementary refinement. Using Llama-3.1-8B-it as an example. Before: using the naive extracted reasoning vectors to steer the model. After: using the complementary refined reasoning vectors to steer the model.}
\label{fig:curve}
\end{figure*}

\subsection{Preliminary Settings}
\label{prelim_setting}





\noindent To investigate whether different types of logical reasoning can be represented linearly in LLMs, we use three datasets-JustLogic, DEER, and ART-corresponding to deductive, inductive, and abductive reasoning, respectively \citep{chen2025justlogic, yang2024language, bhagavatula2019abductive}.
Details about how we use these datasets are provided in Appendix~\ref{sec:ds}. As for the evaluation metric, we choose accuracy for deductive and abductive reasoning and METEOR \cite{banerjee2005meteor} score for inductive reasoning according to the characteristics of different datasets. To ensure fairness, incomplete generations without candidate answers are excluded from accuracy calculations for deductive and abductive reasoning. For inductive reasoning, we take the last three generated sentences, compute METEOR against the ground truth, and use the highest score as the final metric.

For activation extraction and steering, we use layer 13 for both Llama-3.1-8B-it and Gemma-2-9B-it. This choice follows prior work on activation steering, which suggests that middle residual-stream layers tend to contain abstract yet consistently controllable representations, making them suitable intervention points for linear steering directions \citep{rimsky2024steering, zhanguncovering}. Besides, to examine whether the observed pattern extends to larger models with different model structure, we additionally evaluate our method on GPT-OSS-20B \cite{openai2025gptoss120bgptoss20bmodel}, a larger model based on the Mixture-of-Expert (MoE) architecture. More implementation details are provided in Appendix~\ref{sec:id}.

\subsection{Naive Reasoning Vectors Extraction}

Following prior work on contrastive activation analysis and knowledge-vector extraction \citep{rimsky2024steering, zou2023representation, wang2025adaptive}, we firstly construct contrastive activation sets that capture successful versus failed reasoning behaviors. Specifically, we design paired prompts that elicit different levels of reasoning performance: a strong version intended to support correct reasoning and a weak version that leads to failure. During generation, we record the residual stream activations at selected layers for every generated token. For each data instance, if the strong prompt yields a correct reasoning outcome while the weak prompt yields an incorrect one, we treat this as a valid contrastive pair. We then average the recorded activations across the entire generation trajectory to obtain a positive activation vector (successful reasoning) and a negative activation vector (failed reasoning). Formally, each training example provides a pair of activation representations: 

\begin{equation}
\bar{a}_{i,l}^{+} = \frac{1}{N} \sum_{j=1}^{N} a_{j,l}^{+}
\quad \text{and} \quad
\bar{a}_{i,l}^{-} = \frac{1}{N} \sum_{j=1}^{N} a_{j,l}^{-}
\label{eq:avg_pos_neg}
\end{equation}

\noindent where $\bar{a}_{i,l}^{+}$ and $\bar{a}_{i,l}^{-}$ denote the positive (successful reasoning) and negative (failed reasoning) activation vectors for instance $i$ at layer $l$, and the averages are computed over all $N$ token activations collected across the generation process.

Probes help uncover the model’s internal representation mechanisms  \citep{park2023linear, belinkov2022probing, alain2016understanding}. Let $D_{r}$ denote the constructed contrastive activation set for reasoning ${r}$, where ${r}\in\mathcal{P}, \mathcal{P} = \{deductive, inductive, abductive\}$ and $D_{r} = \{\bar{a}_{0,l}^{+}\,...,\bar{a}_{k,l}^{+}\,\bar{a}_{0,l}^{-}\,...,\bar{a}_{k,l}^{-}\}$, we train a linear probe:

\begin{equation}
p_r = \sigma(\theta_{r}^\top x_{i,r} + b_{r}),
\label{eq:probe}
\end{equation}

\noindent where $x_{i,r} \in D_{r}$ denotes the activation sample for reasoning type $r$, and $\theta_{r}$ is the learned reasoning knowledge vector obtained as the probe’s weight parameters. The naive reasoning vector extraction is illustrated in the left panel of Figure~\ref{fig:fig3}.

\subsection{Reasoning Vector Analysis}


As shown in Table~\ref{table:performance}, mono-steering consistently outperforms the unsteered setting, which means the linear assumption works for logical reasoning and these vectors encode corresponding meaningful reasoning knowledge. However, we further investigate the geometric relationships between these vectors by measuring their pairwise cosine similarities and observe that most values are close to zero, as shown in Figure~\ref{fig:fig1}, which means LLMs represent different types of reasoning in largely orthogonal directions \cite{scalena2024multi}. 
In other words, different reasoning knowledge is represented as geometrically distinct directions in the model's activation space. This is contradictory to human cognitive science understanding \citep{holyoak2013oxford, heit2010relations}, which posit that deductive, inductive, and abductive reasoning arise from a common set of cognitive operations and interact in complementary ways during human inference. 

\section{Complementary Reasoning Refinement}

\subsection{Complementary Knowledge Integration}


Motivated by the prior analysis, we propose a complementary subspace-constrained refinement framework, which enables each reasoning vector to learn complementary knowledge from other reasoning types while preserving its own identified features. The framework is illustrated in the right panel of Figure~\ref{fig:fig3}.

\noindent\textbf{Complementary Knowledge Enhancement.} To encourage knowledge sharing across different reasoning types, we introduce the following cosine-based objective:

\begin{equation}
\mathcal{L}_{\text{com}}
= - \sum_{\substack{r,s \in \mathcal{P} \\ r \neq s}}
\frac{\theta_r^\top \theta_s}{\|\theta_r\|_2 \,\|\theta_s\|_2},
\label{eq:cos_loss_sum}
\end{equation}

\noindent where $\theta_r$ and $\theta_s$ denote different reasoning knowledge vectors of type $r$ and $s$. However, relying solely on this complementary loss makes it difficult to control the extent to which the reasoning vectors align. Excessive alignment can cause the vectors to collapse toward a shared direction, leading to the loss of the distinctive features that each vector originally captures.

\noindent\textbf{Reasoning Subspace Constraint.} In order to ensure that complementary knowledge is learned without erasing their distinct characteristics, we introduce a subspace constraint for each reasoning type as well. We note the success of SAEs in extracting fine-grained semantic representations, especially the abstract concepts in LLMs \citep{cunningham2023sparse, hua2025steering, wang2025model}. Here, we use SAEs to construct the reasoning-specific subspace. For each reasoning type $r$, we feed the recorded positive and negative activations into a pretrained SAE to obtain their sparse latent representations:
\begin{equation}
z = \mathrm{SAE\_encode}(h), 
\label{eq:sae_encode}
\end{equation}
where $h$ denotes the residual-stream activation and $z \in \mathbb{R}^m$ is the SAE hidden 
vector with high sparsity.

To identify the SAE features most predictive of reasoning success, we compute the mean squared activation of each latent unit across positive and negative generations:
\begin{equation}
\mu_r^{+}(j) = \mathbb{E}\!\left[z_{j}^{2} \mid \text{pos}\right]
\quad \text{and} \quad
\mu_r^{-}(j) = \mathbb{E}\!\left[z_{j}^{2} \mid \text{neg}\right]
\label{eq:sae_means}
\end{equation}

We then compute a contrastive activation ratio for each SAE feature \cite{muhamed2025saes}:
\begin{equation}
\rho_r(j) = 
\frac{\mu_r^{+}(j)}{\mu_r^{-}(j) + \varepsilon},
\label{eq:sae_ratio}
\end{equation}
where $\varepsilon$ is a small constant ensuring numerical stability. By setting a threshold $\tau$ as the $\alpha$-quantile (e.g., $\alpha=0.9$) of the ratio distribution. Only features with $\rho_r(j) \ge \tau$ are retained as the initial candidate set, ensuring that only 
those units that activate substantially more during successful reasoning than failures are kept. This removes features that are uniformly active or inactive across cases. 

However, a high ratio does not guarantee that the feature contributes substantially in absolute magnitude. Therefore, after ratio-based filtering, we further select the top-$K$ features with the largest mean activation strength, yielding a stable and discriminative reasoning-specific feature set denoted as $\mathcal{F}_r$.

For each feature $j \in \mathcal{F}_r$, we obtain its corresponding decoder direction 
$w^{\mathrm{dec}}_{j}$ from the SAE decoder matrix. Stacking these directions yields a basis
matrix:
\begin{equation}
V_r = \left[ w^{\mathrm{dec}}_{j} \right]_{j \in \mathcal{F}_r} 
\in \mathbb{R}^{d \times K},
\label{eq:decoder_matrix}
\end{equation}
Finally, we orthogonalize this matrix using QR decomposition to obtain a reasoning-specific 
subspace:
\begin{equation}
U_r = \mathrm{QR}(V_r),
\label{eq:qr_orthogonal}
\end{equation}
The columns of $U_r$ form an orthonormal basis capturing the key feature directions associated with reasoning type $r$.

To prevent the refined reasoning vectors from shifting away from their reasoning-specific 
feature structure, we constrain each vector to remain close to its designated subspace. 
Given the orthonormal basis $U_r$ for reasoning type $r$, we decompose the probe vector 
$\theta_r$ into its in-subspace and out-of-subspace components. The orthogonal component 
is penalized through the following subspace-preservation loss:
\begin{equation}
\mathcal{L}_{\text{sub}}^{(r)}
= \left\| (I - U_r U_r^\top)\, \theta_r \right\|_2^2,
\label{eq:sub_loss}
\end{equation}
where $(I - U_rU_r^\top)\theta_r$ extracts the part of $\theta_r$ lying outside the reasoning 
subspace. Minimizing this term ensures that the reasoning vector learns complementary knowledge while retaining the reasoning-specific features encoded by $U_r$.

With the binary cross-entropy loss $\mathcal{L}_{\text{probe}}^{(r)}$ for each reasoning type $r \in \mathcal{P}$, the overall refinement objective is:
\begin{equation}
\mathcal{L} 
= \sum_{r\in\mathcal{P}} \mathcal{L}_{\text{probe}}^{(r)}
\;+\; \lambda_{\text{com}}\,\mathcal{L}_{\text{com}}
\;+\; \lambda_{\text{sub}}\,\sum_{r} \mathcal{L}_{\text{sub}}^{(r)},
\label{eq:total_loss}
\end{equation}
This unified objective jointly encourages: (i) accurate separation of successful and failed 
reasoning behaviors, (ii) controlled sharing of complementary knowledge across reasoning types,
and (iii) preservation of reasoning-specific feature structure within each subspace.

\subsection{Complementary Reasoning Results}

Table~\ref{table:performance} shows that our refined complementary reasoning vectors consistently outperform both the unsteered baseline and the initial reasoning vectors across deductive, inductive, and abductive tasks on Llama-3.1-8B-it and Gemma-2-9B-it, demonstrating more effective and generalizable reasoning representations via SAE-based refinement. As shown in Table~\ref{table:performance}, this trend also generalizes to GPT-OSS-20B, a larger MoE-based model, where both mono steering and complementary steering outperform the unsteered baseline, and complementary steering achieves the strongest overall performance.


It is worth noting that the absolute gains are modest, which is expected given the difficulty of the logical reasoning benchmarks considered in this work. These tasks require multi-step inference and remain challenging even for strong LLMs. We therefore interpret the consistent gains across reasoning types and and model families as evidence that the extracted reasoning vectors capture meaningful aspects of reasoning behavior. The practical significance of our method lies in providing a controllable and analyzable handle on reasoning-related representations.


Figure~\ref{fig:curve} further illustrates the effect of different steering strength on reasoning performance before and after complementary refinement. While performance exhibits a similar dependence on steering strength across all reasoning types, improving at moderate coefficients and degrading under over-steering, our refined vectors achieve higher peak performance at smaller coefficients. This suggests that they capture cleaner and better aligned reasoning directions that require less amplification.

\subsection{Ablation Study}

\begin{table}[t]
\centering
\small
\begin{tabular}{lccc}
\toprule
 & \textbf{Deductive} & \textbf{Inductive} & \textbf{Abductive} \\
\midrule
Our method              & 56.46  & 27.55  & 40.95 \\
\quad w/o (i) & -2.81 & -0.46 & -3.33 \\
\quad w/o (ii) & -4.58 & -0.29 & -2.09 \\
\bottomrule
\end{tabular}

\caption{Ablation study of our complementary method across three logical reasoning types on Llama-3.1-8B-it model. (i)w/o Complementary Konwledge Enhancement (ii) w/o Reasoning Subspace Constraint. }
\label{tab:abl}
\end{table}


Table~\ref{tab:abl} presents the ablation study of the two key components in our complementary refinement: (i) without Complementary Konwledge Enhancement (ii) without Reasoning Subspace Constraint. In each setting, one component is removed at a time. The results show that removing either one of the components leads to performance degradation across all three types of logical reasoning, while combining them together achieves the best performance. This suggests that each type of logical reasoning benefits from auxiliary knowledge transferred from other reasoning types, while unconstrained complementary knowledge integration can negatively impact performance.

\subsection{Cross-Task Generalization on GSM8K}

To evaluate cross-task generalization, we further test the extracted reasoning vectors on GSM8K, a widely used benchmark for multi-step mathematical reasoning. As shown in Table~\ref{tab:gsm8k}, steering with each type of reasoning vector consistently improves over the unsteered baseline, and complementary steering yields further gains over mono steering. Although GSM8K does not isolate a single reasoning type, these results suggest that the learned vectors transfer beyond the original task formulations and retain useful reasoning-related information on an out-of-domain benchmark.

\begin{table}[t]
\centering
\small
\begin{tabular}{lccc}
\toprule
 & \textbf{Deductive} & \textbf{Inductive} & \textbf{Abductive} \\
\midrule
Unsteered              & 76.26  & 76.26  & 76.26 \\
Mono S. & 78.31 & 77.75 & 78.24 \\
Complementary S. & 79.37 & 78.33 & 78.62 \\
\bottomrule
\end{tabular}

\caption{Cross-task generalization on GSM8K. We steer Llama-3.1-8B-it with deductive, inductive, and abductive reasoning vectors, respectively, and report answer accuracy. Across all three vector types, both mono steering and complementary steering improve over the unsteered baseline, with complementary steering achieving the best performance. }
\label{tab:gsm8k}
\end{table}

\section{Analysis}

\subsection{Activated Feature Analysis}


\begin{figure*}[t]
    \centering
    \includegraphics[width=0.8\linewidth]{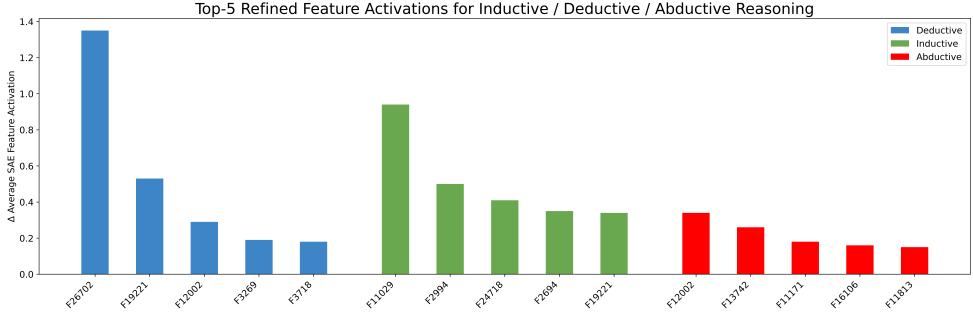}
    \caption{Top-5 refined features for each type of logical reasoning measured by $\Delta_r$ based on SAE.}
    \label{fig:core_f_fig}
\end{figure*}

\begin{figure}[t]
    \centering
    \includegraphics[width=0.8\linewidth]{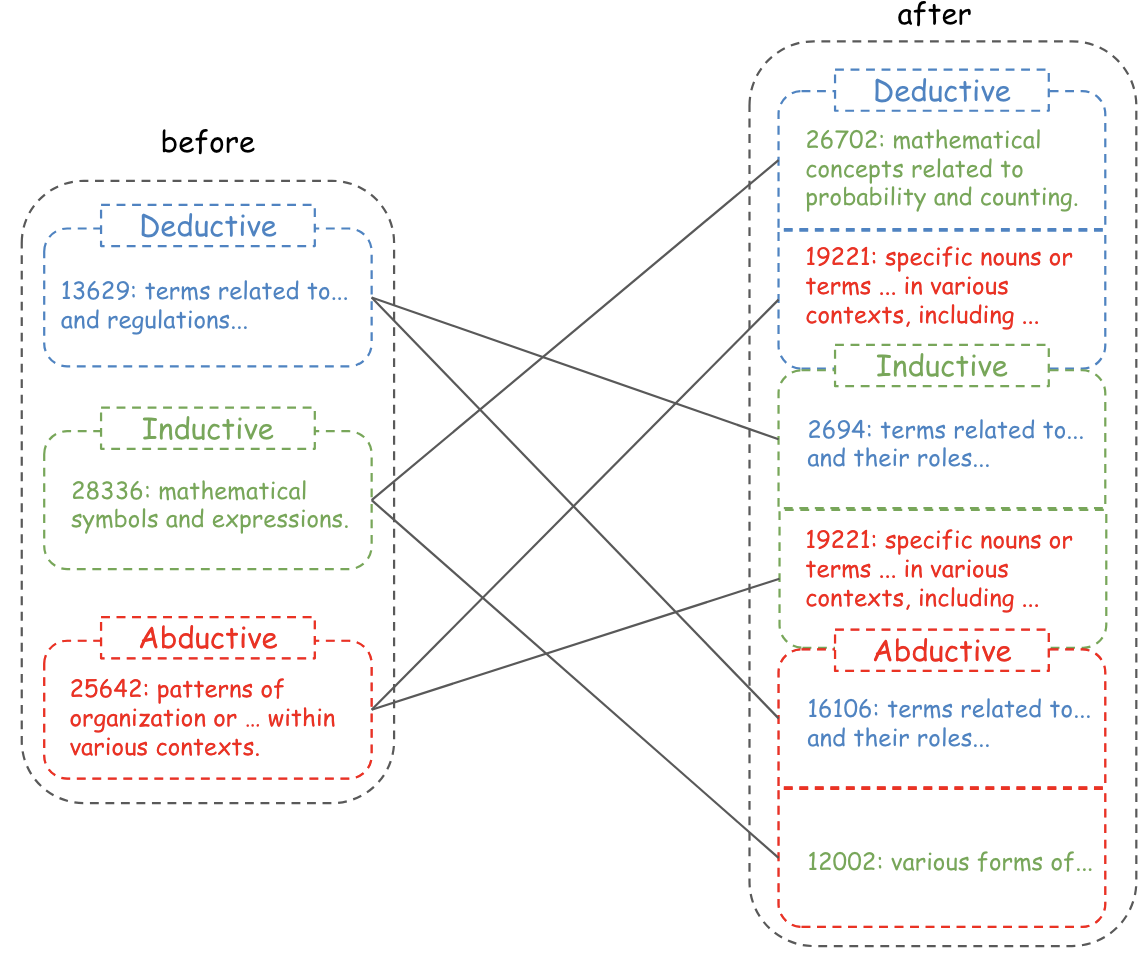}
    \caption{The learned features during complementary refinement process. Take deductive reasoning as an example, $F26702$ and $F19221$ are 2 out of 5 top boosted features after the refinement process from Figure~\ref{fig:core_f_fig}. They have similar semantic explanation with $F28336$ and $F25642$, which are two top features for original inductive and abductive reasoning without refinement.}
    \label{fig:core_exp}
\end{figure}

\begin{figure}[t]
    \centering
    \includegraphics[width=0.8\linewidth]{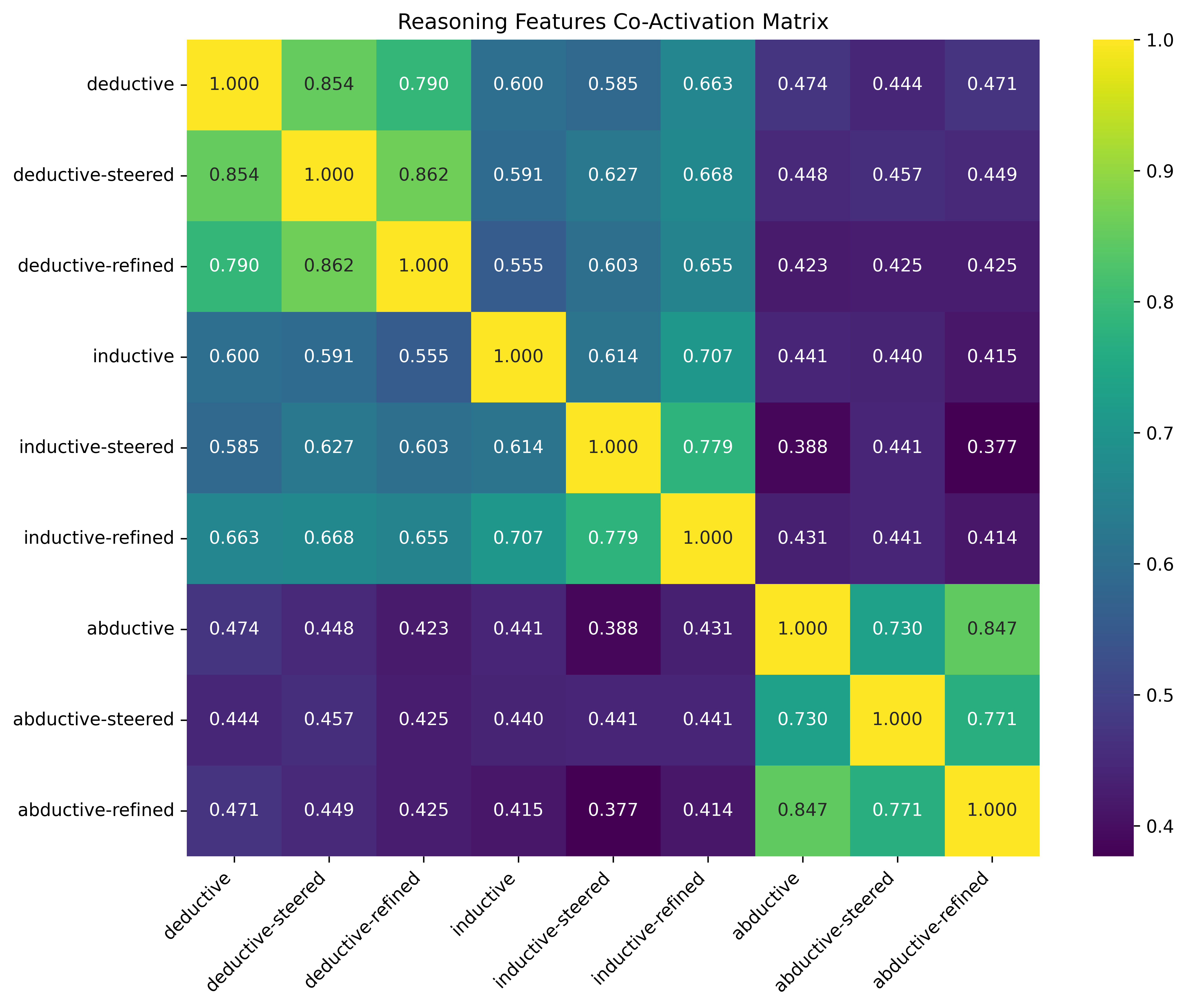}
    \caption{Reasoning Features Co-Activation Matrix.}
    \label{fig:broader_f_fig}
\end{figure}

To understand how reasoning vectors alter internal model representations, we analyze changes in SAE-activated features before and after complementary refinement. Take Llama-3.1-8B-it as an example, we use LlamaScope layer14 residual stream SAE for further fine-grained analysis. We choose layer 14 for analysis because the reasoning vectors are applied at layer 13 to steer model activation (Section \ref{prelim_setting}), so we expect the immediate subsequent residual stream to provide the clearest view of the direct representational changes induced by steering.

\vspace{2pt}

\noindent\textbf{Core features Analysis.} Given a set of residual stream activations collected from layer 14, we encode each activation sequence into the SAE latent space following Equation (4). For each reasoning type, we compute the mean SAE activation vectors before and after refinement and obtain their feature-wise difference:

\[
\Delta_r = \bar{z}^{\text{ref}}_r - \bar{z}^{\text{orig}}_r
\] 

Then, we identify the Top-5 SAE features exhibiting the largest difference in activation for each reasoning type. These features are visualized in Figure~\ref{fig:core_f_fig} to illustrate how refinement shifts the underlying semantic subspaces associated with deductive, inductive, and abductive reasoning. We use Neuronpedia \cite{lin2023neuronpedia} to retrieve the feature explanations. For each type of logical reasoning, we observe that the top boosted features share similar semantic relevance to the top features activated in the other two reasoning types. This indicates that our complementary refinement enables each reasoning type to effectively leverage meaningful features learned from the others. The related feature explanations are shown in Figure~\ref{fig:core_exp}.

\vspace{2pt}

\noindent\textbf{Broader features Analysis.} While the core-feature analysis focuses on the features most strongly affected by refinement, we additionally investigate how the broader distribution of reasoning-related SAE feature activations changes as the steering or refined-steering intervention added. For each reasoning dataset and model variant (unsteered, mono-steered, and refined-steered), we encode all residual stream activations from layer 14 following Equation (4) and compute the mean activation vector for each variant setting as $x$.
To quantify how similarly two variant settings activate their respective top-k SAE features (k=100 in our experiments), we compute the symmetric co-activation similarity between variant settings i and j as:

\[
S(i,j) = \frac{2 \sum_{k} \min(x_i(k),\, x_j(k))}
              {\sum_{k} x_i(k) + \sum_{k} x_j(k)}
\]

This measures how strongly the two top-feature sets align in expectation over the SAE space; a value near 1 indicates heavy overlap in activated feature directions, while lower values indicate divergence. We compute this similarity across all nine variant settings (three reasoning types times three model versions), producing a 9×9 co-activation matrix visualized in Figure~\ref{fig:broader_f_fig}.

As the result shows to us, deductive and inductive reasoning become more aligned in their overall activation distributions, whereas abductive reasoning becomes more differentiated. This trend is reflected in the co-activation scores: on one hand, it is apparent that the deductive–inductive reasoning region displays higher co-activation values; on the other hand, from a fine-grained perspective, deductive–inductive similarity rises from 0.600 to 0.655 across variants, while abductive–deductive similarity (0.474 → 0.457 → 0.425) and abductive–inductive (0.441 → 0.441 → 0.414) similarities consistently decline. These patterns are consistent with the fact that deductive and inductive reasoning share an evidence-based inferential structure, whereas abductive reasoning relies more on explanation-based inference.

Former core feature analysis shows that the Top-5 feature shifts of each refined reasoning vector often include semantic cues associated with other reasoning types. This does not contradict the broader divergence observed above because LLM representations are compositional: complementary refinement shapes global reasoning structure while still allowing shared cross-reasoning features that supports complementary reasoning behaviors.


\subsection{Reasoning Causal Tracing}

In order to better understand the impact of the naive and complementary refined reasoning knowledge vectors, besides the feature level analysis, we further conduct mechanism analysis in models' activation space to trace the differentiated causal traits before and after our complementary refinement.

Activation patching is a mechanistic interpretability technique that measures the causal contribution of specific model components by replacing their activations and observing the resulting change in model behavior \citep{zhang2023towards, conmy2023towards}. Here, we following \citet{heimersheim2024use}, which overwrite specific activations during a model run with cached activations from a previous run and observe how this affects the model’s output. We focus on patching the activations of attention heads. Detailed operation and results are provided in Appendix~\ref{sec:ap}. We have following observations from Figure~\ref{fig:att_patching}:



\noindent \textbf{Core activations reserved.} The attention heads that were strongly aligned with current reasoning remain consistently active after complementary refinement, such as layer 27 head 29 for abductive reasoning and layer 17 head 24 for inductive reasoning. This persistence indicates that the refinement procedure preserves the core representational components necessary for the targeted reasoning behavior. Notably, certain heads such as layer 31 head 14 remain relatively highly-activated across all three reasoning types, suggesting their role as general-purpose reasoning facilitators.

\noindent \textbf{Activation profile concentrated.} The refined-steered model exhibits a more concentrated activation profile: the most influential heads become more localized, while diffuse non-essential activations diminish. In several cases, such as layer 16 head 30 and layer 31 head 14 for inductive reasoning, refinement increases activation strength, implying improved sparsity, specialization, and more efficient utilization of reasoning-relevant circuits.

\noindent \textbf{New activations appear.} Refinement gives rise to newly activated attention heads that are not prominent in the naive-steered model. For example, layer 17 head 24 becomes salient for deductive reasoning after refinement, while layer 24 head 27 shows increased activation for both deductive and abductive reasoning. These new heads suggests that refinement not only preserves existing reasoning circuits but also recruits additional components that encode complementary reasoning features.


\subsection{Qualitative Analysis}





\begin{table}[t]
\centering
\small

\begin{tabular}{p{7cm}}
\toprule

...Since both conditions support the statement, we \hlD{can conclude that the statement} is true. \\
\midrule

...is equivalent to saying "buffalo do not roam the pr\hlD{airie." Since the} paragraph states that the notion "buffalo roam the prairie" can be considered false,... \\
\midrule

...but does not mention insecticides or industries that \hlD{sell them. Therefore,} the truth value of the statement is uncertain.
 \\
\midrule

...has three main body parts - head, thorax, and abdomen, then it has \hlI{three pairs of legs.} \\
\midrule

...is the deepest part of the ocean and the deepest location on Earth. It is 11,034 meters (36,201 feet) deep, \hlI{which is almost 7} miles... \\
\midrule

...can have two types of sales assistants: those that help the \hlI{customer and those that} keep the space and clothes organized...\\
\midrule

...Hypothesis B "Our founder Rachel only uses the PC" seems \hlA{more plausible because it is} possible that...
 \\
\midrule

...have time to wash the existing ones\hlA{. Hypothesis} A, which states that Lina received a pile of adult clothes, seems... \\
\midrule

...has to close again tomorrow. This implies that his work schedule \hlA{is likely a result of} his preference for... \\

\bottomrule
\end{tabular}

\caption{Representative text spans extracted for \hlD{deductive}, \hlI{inductive} and \hlA{abductive} reasoning.}
\label{tab:highlight_examples}
\end{table}

Since prior work reveals that causal connectives such as 'therefore', 'thus', and 'because' explicitly encode inferential relations in text \cite{sanders2012causal}, we investigate whether meaningful knowledge in reasoning vectors actually highlighted specific keywords or phrases in text distributions. Following the text span extraction procedure in \citet{han2024word}, we quantify the influence of a reasoning vector by computing the log-likelihood shift for each token by $\Delta_i = \log P_{r}(v_i \mid v_{<i}) - \log P_0(v_i \mid v_{<i})$ and apply a dynamic programming algorithm to identify the contiguous token span with the highest cumulative shift under length 5, yielding a concise summary of the textual patterns most affected by the reasoning vector. Several extracted representative spans are presented in Table~\ref{tab:highlight_examples}. Deductive reasoning vectors tend to emphasize causal connectives such as 'therefore', 'since' and explicit conclusion markers, which are characteristic of rule-based logical reasoning. Inductive reasoning vectors frequently attend to generalization-related phrases, including quantifiers, categorical properties, and statistical regularities, reflecting reasoning from repeated observations. In contrast, abductive reasoning vectors highlight plausibility-oriented expressions such as 'more plausible' and 'likely', which are indicative of hypothesis selection under uncertainty. These findings suggest that the learned reasoning vectors capture semantically meaningful linguistic cues associated with distinct reasoning processes, providing qualitative evidence that the vectors encode interpretable reasoning-related knowledge rather than arbitrary activation patterns.

\section{Related Work}

\textbf{Logical Reasoning in LLMs.} Logical reasoning includes multiple forms of inference that enable humans and machines to draw conclusions from some premises or observations \cite{johnson2010mental}. Classical cognitive science distinguishes three primary types of reasoning: deductive, inductive, and abductive \cite{peirce1934collected}, which have increasingly been adopted as frameworks for analyzing the behavior of LLMs \cite{li2025patterns}. Several works enhance reasoning via strategy selection or hybrid deductive–inductive designs \citep{cheng2024inductivedeductiverethinkingfundamental, cai2025role, wang2024typedthinker}. However, most of current work focus on external reasoning paths, with limited analysis of internal representations in activation space \citep{tan2024analysing, dumas2025separating}. In contrast, we study whether logical reasoning can be linearly manipulated to enable controllable and compositional reasoning behaviors \citep{scalena2024multi, nguyen2025multi}.

\noindent\textbf{Knowledge Vector.} Knowledge vectors (KVs) in LLMs refer to linear representations that encode semantic information in model parameters or activation space \cite{elhage2022toy}. Early work on model editing shows that factual knowledge can be localized and manipulated via linear interventions \citep{meng2022locating, meng2022mass}, and more recent studies identify latent directions corresponding to high-level behaviors such as sentiment and instruction-following \citep{turner2023steering, stolfo2024improving}. Despite these advances, reasoning-related vectors remain largely unexplored. Compared to prior work that focuses on isolated behaviors such as chain-of-thought stages or backtracking patterns \citep{venhoff2025understanding, zhanguncovering}, we investigate whether general logical reasoning types correspond to separable directions in activation space.


\section{Conclusion}

In this work, we show that logical reasoning can be linearly represented as distinct knowledge vectors in LLM activation space and geometric analysis reveals low pairwise cosine similarity among these vectors, suggesting that different reasoning abilities rely on distinct representational mechanisms. Motivated by above observation, we hypothesize that reasoning vectors can be improved by incorporating complementary knowledge across reasoning types and propose a complementary subspace–constrained refinement framework based on SAEs, which preserves each vector’s core features while selectively integrating complementary information. Experiments show that the refined vectors consistently outperform both the unsteered and mono-steering baselines, and further multi-perspective analyses reveals the shared and specific features among these reasoning knowledge vectors.

\section*{Limitations}

One limitation of this work is that our study focuses on three logical  reasoning datasets. Our experiments do not explore how the proposed approach generalizes to other reasoning benchmarks, nor do we examine potential cross-domain transfer effects between datasets. In addition, we do not investigate more fine-grained reasoning forms, such as analogical reasoning and counterfactual reasoning, which may involve different representational structures or interaction patterns in model's activation space.


\bibliography{custom}

@misc{openai2024gpt4technicalreport,
      title={GPT-4 Technical Report}, 
      author={OpenAI and Josh Achiam and Steven Adler and Sandhini Agarwal and Lama Ahmad and Ilge Akkaya and Florencia Leoni Aleman and Diogo Almeida and Janko Altenschmidt and Sam Altman and Shyamal Anadkat and Red Avila and Igor Babuschkin and Suchir Balaji and Valerie Balcom and Paul Baltescu and Haiming Bao and Mohammad Bavarian and Jeff Belgum and Irwan Bello and Jake Berdine and Gabriel Bernadett-Shapiro and Christopher Berner and Lenny Bogdonoff and Oleg Boiko and Madelaine Boyd and Anna-Luisa Brakman and Greg Brockman and Tim Brooks and Miles Brundage and Kevin Button and Trevor Cai and Rosie Campbell and Andrew Cann and Brittany Carey and Chelsea Carlson and Rory Carmichael and Brooke Chan and Che Chang and Fotis Chantzis and Derek Chen and Sully Chen and Ruby Chen and Jason Chen and Mark Chen and Ben Chess and Chester Cho and Casey Chu and Hyung Won Chung and Dave Cummings and Jeremiah Currier and Yunxing Dai and Cory Decareaux and Thomas Degry and Noah Deutsch and Damien Deville and Arka Dhar and David Dohan and Steve Dowling and Sheila Dunning and Adrien Ecoffet and Atty Eleti and Tyna Eloundou and David Farhi and Liam Fedus and Niko Felix and Simón Posada Fishman and Juston Forte and Isabella Fulford and Leo Gao and Elie Georges and Christian Gibson and Vik Goel and Tarun Gogineni and Gabriel Goh and Rapha Gontijo-Lopes and Jonathan Gordon and Morgan Grafstein and Scott Gray and Ryan Greene and Joshua Gross and Shixiang Shane Gu and Yufei Guo and Chris Hallacy and Jesse Han and Jeff Harris and Yuchen He and Mike Heaton and Johannes Heidecke and Chris Hesse and Alan Hickey and Wade Hickey and Peter Hoeschele and Brandon Houghton and Kenny Hsu and Shengli Hu and Xin Hu and Joost Huizinga and Shantanu Jain and Shawn Jain and Joanne Jang and Angela Jiang and Roger Jiang and Haozhun Jin and Denny Jin and Shino Jomoto and Billie Jonn and Heewoo Jun and Tomer Kaftan and Łukasz Kaiser and Ali Kamali and Ingmar Kanitscheider and Nitish Shirish Keskar and Tabarak Khan and Logan Kilpatrick and Jong Wook Kim and Christina Kim and Yongjik Kim and Jan Hendrik Kirchner and Jamie Kiros and Matt Knight and Daniel Kokotajlo and Łukasz Kondraciuk and Andrew Kondrich and Aris Konstantinidis and Kyle Kosic and Gretchen Krueger and Vishal Kuo and Michael Lampe and Ikai Lan and Teddy Lee and Jan Leike and Jade Leung and Daniel Levy and Chak Ming Li and Rachel Lim and Molly Lin and Stephanie Lin and Mateusz Litwin and Theresa Lopez and Ryan Lowe and Patricia Lue and Anna Makanju and Kim Malfacini and Sam Manning and Todor Markov and Yaniv Markovski and Bianca Martin and Katie Mayer and Andrew Mayne and Bob McGrew and Scott Mayer McKinney and Christine McLeavey and Paul McMillan and Jake McNeil and David Medina and Aalok Mehta and Jacob Menick and Luke Metz and Andrey Mishchenko and Pamela Mishkin and Vinnie Monaco and Evan Morikawa and Daniel Mossing and Tong Mu and Mira Murati and Oleg Murk and David Mély and Ashvin Nair and Reiichiro Nakano and Rajeev Nayak and Arvind Neelakantan and Richard Ngo and Hyeonwoo Noh and Long Ouyang and Cullen O'Keefe and Jakub Pachocki and Alex Paino and Joe Palermo and Ashley Pantuliano and Giambattista Parascandolo and Joel Parish and Emy Parparita and Alex Passos and Mikhail Pavlov and Andrew Peng and Adam Perelman and Filipe de Avila Belbute Peres and Michael Petrov and Henrique Ponde de Oliveira Pinto and Michael and Pokorny and Michelle Pokrass and Vitchyr H. Pong and Tolly Powell and Alethea Power and Boris Power and Elizabeth Proehl and Raul Puri and Alec Radford and Jack Rae and Aditya Ramesh and Cameron Raymond and Francis Real and Kendra Rimbach and Carl Ross and Bob Rotsted and Henri Roussez and Nick Ryder and Mario Saltarelli and Ted Sanders and Shibani Santurkar and Girish Sastry and Heather Schmidt and David Schnurr and John Schulman and Daniel Selsam and Kyla Sheppard and Toki Sherbakov and Jessica Shieh and Sarah Shoker and Pranav Shyam and Szymon Sidor and Eric Sigler and Maddie Simens and Jordan Sitkin and Katarina Slama and Ian Sohl and Benjamin Sokolowsky and Yang Song and Natalie Staudacher and Felipe Petroski Such and Natalie Summers and Ilya Sutskever and Jie Tang and Nikolas Tezak and Madeleine B. Thompson and Phil Tillet and Amin Tootoonchian and Elizabeth Tseng and Preston Tuggle and Nick Turley and Jerry Tworek and Juan Felipe Cerón Uribe and Andrea Vallone and Arun Vijayvergiya and Chelsea Voss and Carroll Wainwright and Justin Jay Wang and Alvin Wang and Ben Wang and Jonathan Ward and Jason Wei and CJ Weinmann and Akila Welihinda and Peter Welinder and Jiayi Weng and Lilian Weng and Matt Wiethoff and Dave Willner and Clemens Winter and Samuel Wolrich and Hannah Wong and Lauren Workman and Sherwin Wu and Jeff Wu and Michael Wu and Kai Xiao and Tao Xu and Sarah Yoo and Kevin Yu and Qiming Yuan and Wojciech Zaremba and Rowan Zellers and Chong Zhang and Marvin Zhang and Shengjia Zhao and Tianhao Zheng and Juntang Zhuang and William Zhuk and Barret Zoph},
      year={2024},
      eprint={2303.08774},
      archivePrefix={arXiv},
      primaryClass={cs.CL},
      url={https://arxiv.org/abs/2303.08774}, 
}

@inproceedings{
guan2023leveraging,
title={Leveraging Pre-trained Large Language Models to Construct and Utilize World Models for Model-based Task Planning},
author={Lin Guan and Karthik Valmeekam and Sarath Sreedharan and Subbarao Kambhampati},
booktitle={Thirty-seventh Conference on Neural Information Processing Systems},
year={2023},
url={https://openreview.net/forum?id=zDbsSscmuj}
}

@inproceedings{yao2022react,
  title={React: Synergizing reasoning and acting in language models},
  author={Yao, Shunyu and Zhao, Jeffrey and Yu, Dian and Du, Nan and Shafran, Izhak and Narasimhan, Karthik R and Cao, Yuan},
  booktitle={The eleventh international conference on learning representations},
  year={2022}
}

@inproceedings{pan2023logic,
  title={Logic-lm: Empowering large language models with symbolic solvers for faithful logical reasoning},
  author={Pan, Liangming and Albalak, Alon and Wang, Xinyi and Wang, William},
  booktitle={Findings of the Association for Computational Linguistics: EMNLP 2023},
  pages={3806--3824},
  year={2023}
}

@inproceedings{lam2024closer,
  title={A Closer Look at Tool-based Logical Reasoning with LLMs: The Choice of Tool Matters},
  author={Lam, Long Hei Matthew and Thatikonda, Ramya Keerthy and Shareghi, Ehsan},
  booktitle={Proceedings of the 22nd Annual Workshop of the Australasian Language Technology Association},
  pages={41--63},
  year={2024}
}

@book{peirce1934collected,
  title={Collected papers of charles sanders peirce},
  author={Peirce, Charles Sanders},
  volume={5},
  year={1934},
  publisher={Harvard University Press}
}

@inproceedings{kang2025exploring,
  title={Exploring Deductive and Inductive Reasoning Capabilities of Large Language Models in Procedural Planning},
  author={Kang, Jiabao and Li, Xinye and Xu, Liyan and Liu, Qingbin and Chen, Xi and Tu, Zhiying and Chu, Dianhui and Sui, Dianbo},
  booktitle={Findings of the Association for Computational Linguistics: EMNLP 2025},
  pages={16320--16341},
  year={2025}
}

@article{vashishtha2025executable,
  title={Executable Counterfactuals: Improving LLMs' Causal Reasoning Through Code},
  author={Vashishtha, Aniket and Dai, Qirun and Mei, Hongyuan and Sharma, Amit and Tan, Chenhao and Peng, Hao},
  journal={arXiv preprint arXiv:2510.01539},
  year={2025}
}

@article{abdaljalil2025theorem,
  title={Theorem-of-Thought: A Multi-Agent Framework for Abductive, Deductive, and Inductive Reasoning in Language Models},
  author={Abdaljalil, Samir and Kurban, Hasan and Qaraqe, Khalid and Serpedin, Erchin},
  journal={arXiv preprint arXiv:2506.07106},
  year={2025}
}

@article{bedi2025fidelity,
  title={Fidelity of medical reasoning in large language models},
  author={Bedi, Suhana and Jiang, Yixing and Chung, Philip and Koyejo, Sanmi and Shah, Nigam},
  journal={JAMA Network Open},
  volume={8},
  number={8},
  pages={e2526021--e2526021},
  year={2025},
  publisher={American Medical Association}
}

@inproceedings{rimsky2024steering,
  title={Steering llama 2 via contrastive activation addition},
  author={Rimsky, Nina and Gabrieli, Nick and Schulz, Julian and Tong, Meg and Hubinger, Evan and Turner, Alexander},
  booktitle={Proceedings of the 62nd Annual Meeting of the Association for Computational Linguistics (Volume 1: Long Papers)},
  pages={15504--15522},
  year={2024}
}

@article{turner2023steering,
  title={Steering language models with activation engineering},
  author={Turner, Alexander Matt and Thiergart, Lisa and Leech, Gavin and Udell, David and Vazquez, Juan J and Mini, Ulisse and MacDiarmid, Monte},
  journal={arXiv preprint arXiv:2308.10248},
  year={2023}
}

@article{jorgensen2023improving,
  title={Improving activation steering in language models with mean-centring},
  author={Jorgensen, Ole and Cope, Dylan and Schoots, Nandi and Shanahan, Murray},
  journal={arXiv preprint arXiv:2312.03813},
  year={2023}
}

@article{zou2023representation,
  title={Representation engineering: A top-down approach to ai transparency},
  author={Zou, Andy and Phan, Long and Chen, Sarah and Campbell, James and Guo, Phillip and Ren, Richard and Pan, Alexander and Yin, Xuwang and Mazeika, Mantas and Dombrowski, Ann-Kathrin and others},
  journal={arXiv preprint arXiv:2310.01405},
  year={2023}
}

@article{nguyen2025multi,
  title={Multi-attribute steering of language models via targeted intervention},
  author={Nguyen, Duy and Prasad, Archiki and Stengel-Eskin, Elias and Bansal, Mohit},
  journal={arXiv preprint arXiv:2502.12446},
  year={2025}
}

@inproceedings{wang2025adaptive,
  title={Adaptive activation steering: A tuning-free llm truthfulness improvement method for diverse hallucinations categories},
  author={Wang, Tianlong and Jiao, Xianfeng and Zhu, Yinghao and Chen, Zhongzhi and He, Yifan and Chu, Xu and Gao, Junyi and Wang, Yasha and Ma, Liantao},
  booktitle={Proceedings of the ACM on Web Conference 2025},
  pages={2562--2578},
  year={2025}
}

@article{stolfo2024improving,
  title={Improving instruction-following in language models through activation steering},
  author={Stolfo, Alessandro and Balachandran, Vidhisha and Yousefi, Safoora and Horvitz, Eric and Nushi, Besmira},
  journal={arXiv preprint arXiv:2410.12877},
  year={2024}
}

@article{venhoff2025understanding,
  title={Understanding reasoning in thinking language models via steering vectors},
  author={Venhoff, Constantin and Arcuschin, Iv{\'a}n and Torr, Philip and Conmy, Arthur and Nanda, Neel},
  journal={arXiv preprint arXiv:2506.18167},
  year={2025}
}

@inproceedings{zhanguncovering,
  title={Uncovering Latent Chain of Thought Vectors in Large Language Models},
  author={Zhang, Jason and Viteri, Scott W},
  booktitle={Workshop on Neural Network Weights as a New Data Modality}
}

@article{scalena2024multi,
  title={Multi-property steering of large language models with dynamic activation composition},
  author={Scalena, Daniel and Sarti, Gabriele and Nissim, Malvina},
  journal={arXiv preprint arXiv:2406.17563},
  year={2024}
}

@book{holyoak2013oxford,
  title={The Oxford handbook of thinking and reasoning},
  author={Holyoak, Keith J and Morrison, Robert G},
  year={2013},
  publisher={Oxford University Press}
}

@inproceedings{cai2025role,
  title={The role of deductive and inductive reasoning in large language models},
  author={Cai, Chengkun and Zhao, Xu and Liu, Haoliang and Jiang, Zhongyu and Zhang, Tianfang and Wu, Zongkai and Hwang, Jenq-Neng and Li, Lei},
  booktitle={Proceedings of the 63rd Annual Meeting of the Association for Computational Linguistics (Volume 1: Long Papers)},
  pages={16780--16790},
  year={2025}
}

@misc{nanda2023othello,
  author       = {Nanda, Neel},
  title        = {Actually, Othello-GPT Has a Linear Emergent World Representation},
  howpublished = {\url{https://www.neelnanda.io/mechanistic-interpretability/othello}},
  year         = {2023}
}

@article{shu2025survey,
  title={A survey on sparse autoencoders: Interpreting the internal mechanisms of large language models},
  author={Shu, Dong and Wu, Xuansheng and Zhao, Haiyan and Rai, Daking and Yao, Ziyu and Liu, Ninghao and Du, Mengnan},
  journal={arXiv preprint arXiv:2503.05613},
  year={2025}
}

@article{heimersheim2024use,
  title={How to use and interpret activation patching},
  author={Heimersheim, Stefan and Nanda, Neel},
  journal={arXiv preprint arXiv:2404.15255},
  year={2024}
}

@article{zhang2023towards,
  title={Towards best practices of activation patching in language models: Metrics and methods},
  author={Zhang, Fred and Nanda, Neel},
  journal={arXiv preprint arXiv:2309.16042},
  year={2023}
}

@inproceedings{han2024word,
  title={Word embeddings are steers for language models},
  author={Han, Chi and Xu, Jialiang and Li, Manling and Fung, Yi and Sun, Chenkai and Jiang, Nan and Abdelzaher, Tarek and Ji, Heng},
  booktitle={Proceedings of the 62nd Annual Meeting of the Association for Computational Linguistics (Volume 1: Long Papers)},
  pages={16410--16430},
  year={2024}
}

@article{johnson2010mental,
  title={Mental models and human reasoning},
  author={Johnson-Laird, Philip N},
  journal={Proceedings of the National Academy of Sciences},
  volume={107},
  number={43},
  pages={18243--18250},
  year={2010},
  publisher={National Academy of Sciences}
}

@article{li2025patterns,
  title={Patterns over principles: The fragility of inductive reasoning in llms under noisy observations},
  author={Li, Chunyang and Wang, Weiqi and Zheng, Tianshi and Song, Yangqiu},
  journal={arXiv preprint arXiv:2502.16169},
  year={2025}
}

@misc{cheng2024inductivedeductiverethinkingfundamental,
      title={Inductive or Deductive? Rethinking the Fundamental Reasoning Abilities of LLMs}, 
      author={Kewei Cheng and Jingfeng Yang and Haoming Jiang and Zhengyang Wang and Binxuan Huang and Ruirui Li and Shiyang Li and Zheng Li and Yifan Gao and Xian Li and Bing Yin and Yizhou Sun},
      year={2024},
      eprint={2408.00114},
      archivePrefix={arXiv},
      primaryClass={cs.AI},
      url={https://arxiv.org/abs/2408.00114}, 
}

@article{wang2024typedthinker,
  title={TypedThinker: Diversify Large Language Model Reasoning with Typed Thinking},
  author={Wang, Danqing and Ma, Jianxin and Fang, Fei and Li, Lei},
  journal={arXiv preprint arXiv:2410.01952},
  year={2024}
}

@article{tan2024analysing,
  title={Analysing the generalisation and reliability of steering vectors},
  author={Tan, Daniel and Chanin, David and Lynch, Aengus and Paige, Brooks and Kanoulas, Dimitrios and Garriga-Alonso, Adri{\`a} and Kirk, Robert},
  journal={Advances in Neural Information Processing Systems},
  volume={37},
  pages={139179--139212},
  year={2024}
}

@inproceedings{dumas2025separating,
  title={Separating tongue from thought: Activation patching reveals language-agnostic concept representations in transformers},
  author={Dumas, Cl{\'e}ment and Wendler, Chris and Veselovsky, Veniamin and Monea, Giovanni and West, Robert},
  booktitle={Proceedings of the 63rd Annual Meeting of the Association for Computational Linguistics (Volume 1: Long Papers)},
  pages={31822--31841},
  year={2025}
}

@article{meng2022locating,
  title={Locating and editing factual associations in gpt},
  author={Meng, Kevin and Bau, David and Andonian, Alex and Belinkov, Yonatan},
  journal={Advances in neural information processing systems},
  volume={35},
  pages={17359--17372},
  year={2022}
}

@article{meng2022mass,
  title={Mass-editing memory in a transformer},
  author={Meng, Kevin and Sharma, Arnab Sen and Andonian, Alex and Belinkov, Yonatan and Bau, David},
  journal={arXiv preprint arXiv:2210.07229},
  year={2022}
}

@article{elhage2022toy,
  title={Toy models of superposition},
  author={Elhage, Nelson and Hume, Tristan and Olsson, Catherine and Schiefer, Nicholas and Henighan, Tom and Kravec, Shauna and Hatfield-Dodds, Zac and Lasenby, Robert and Drain, Dawn and Chen, Carol and others},
  journal={arXiv preprint arXiv:2209.10652},
  year={2022}
}

@article{park2023linear,
  title={The linear representation hypothesis and the geometry of large language models},
  author={Park, Kiho and Choe, Yo Joong and Veitch, Victor},
  journal={arXiv preprint arXiv:2311.03658},
  year={2023}
}

@article{belinkov2022probing,
  title={Probing classifiers: Promises, shortcomings, and advances},
  author={Belinkov, Yonatan},
  journal={Computational Linguistics},
  volume={48},
  number={1},
  pages={207--219},
  year={2022}
}

@article{alain2016understanding,
  title={Understanding intermediate layers using linear classifier probes},
  author={Alain, Guillaume and Bengio, Yoshua},
  journal={arXiv preprint arXiv:1610.01644},
  year={2016}
}

@article{cunningham2023sparse,
  title={Sparse autoencoders find highly interpretable features in language models},
  author={Cunningham, Hoagy and Ewart, Aidan and Riggs, Logan and Huben, Robert and Sharkey, Lee},
  journal={arXiv preprint arXiv:2309.08600},
  year={2023}
}

@article{hua2025steering,
  title={Steering LVLMs via Sparse Autoencoder for Hallucination Mitigation},
  author={Hua, Zhenglin and He, Jinghan and Yao, Zijun and Han, Tianxu and Guo, Haiyun and Jia, Yuheng and Fang, Junfeng},
  journal={arXiv preprint arXiv:2505.16146},
  year={2025}
}

@article{wang2025model,
  title={Model Unlearning via Sparse Autoencoder Subspace Guided Projections},
  author={Wang, Xu and Li, Zihao and Wang, Benyou and Hu, Yan and Zou, Difan},
  journal={arXiv preprint arXiv:2505.24428},
  year={2025}
}

@inproceedings{muhamed2025saes,
  title={Saes can improve unlearning: Dynamic sparse autoencoder guardrails for precision unlearning in llms},
  author={Muhamed, Aashiq and Bonato, Jacopo and Diab, Mona T and Smith, Virginia},
  booktitle={ICML 2025 Workshop on Reliable and Responsible Foundation Models},
  year={2025}
}

@article{chen2025justlogic,
  title={Justlogic: A comprehensive benchmark for evaluating deductive reasoning in large language models},
  author={Chen, Michael K and Zhang, Xikun and Tao, Dacheng},
  journal={arXiv preprint arXiv:2501.14851},
  year={2025}
}

@inproceedings{yang2024language,
  title={Language models as inductive reasoners},
  author={Yang, Zonglin and Dong, Li and Du, Xinya and Cheng, Hao and Cambria, Erik and Liu, Xiaodong and Gao, Jianfeng and Wei, Furu},
  booktitle={Proceedings of the 18th Conference of the European Chapter of the Association for Computational Linguistics (Volume 1: Long Papers)},
  pages={209--225},
  year={2024}
}

@article{bhagavatula2019abductive,
  title={Abductive commonsense reasoning},
  author={Bhagavatula, Chandra and Bras, Ronan Le and Malaviya, Chaitanya and Sakaguchi, Keisuke and Holtzman, Ari and Rashkin, Hannah and Downey, Doug and Yih, Scott Wen-tau and Choi, Yejin},
  journal={arXiv preprint arXiv:1908.05739},
  year={2019}
}

@article{grattafiori2024llama,
  title={The llama 3 herd of models},
  author={Grattafiori, Aaron and Dubey, Abhimanyu and Jauhri, Abhinav and Pandey, Abhinav and Kadian, Abhishek and Al-Dahle, Ahmad and Letman, Aiesha and Mathur, Akhil and Schelten, Alan and Vaughan, Alex and others},
  journal={arXiv preprint arXiv:2407.21783},
  year={2024}
}

@article{team2024gemma,
  title={Gemma 2: Improving open language models at a practical size},
  author={Team, Gemma and Riviere, Morgane and Pathak, Shreya and Sessa, Pier Giuseppe and Hardin, Cassidy and Bhupatiraju, Surya and Hussenot, L{\'e}onard and Mesnard, Thomas and Shahriari, Bobak and Ram{\'e}, Alexandre and others},
  journal={arXiv preprint arXiv:2408.00118},
  year={2024}
}

@misc{kissane2024sae,
  title        = {{SAEs (usually) Transfer Between Base and Chat Models}},
  author       = {Kissane, Connor and robertzk and Conmy, Arthur and Nanda, Neel},
  howpublished = {\url{https://www.lesswrong.com/posts/fmwk6qxrpW8d4jvbd/saes-usually-transfer-between-base-and-chat-models}},
  year         = {2024},
  month        = {July}
}

@article{lieberum2024gemma,
  title={Gemma scope: Open sparse autoencoders everywhere all at once on gemma 2},
  author={Lieberum, Tom and Rajamanoharan, Senthooran and Conmy, Arthur and Smith, Lewis and Sonnerat, Nicolas and Varma, Vikrant and Kram{\'a}r, J{\'a}nos and Dragan, Anca and Shah, Rohin and Nanda, Neel},
  journal={arXiv preprint arXiv:2408.05147},
  year={2024}
}

@article{he2024llama,
  title={Llama scope: Extracting millions of features from llama-3.1-8b with sparse autoencoders},
  author={He, Zhengfu and Shu, Wentao and Ge, Xuyang and Chen, Lingjie and Wang, Junxuan and Zhou, Yunhua and Liu, Frances and Guo, Qipeng and Huang, Xuanjing and Wu, Zuxuan and others},
  journal={arXiv preprint arXiv:2410.20526},
  year={2024}
}

@inproceedings{banerjee2005meteor,
  title={METEOR: An automatic metric for MT evaluation with improved correlation with human judgments},
  author={Banerjee, Satanjeev and Lavie, Alon},
  booktitle={Proceedings of the acl workshop on intrinsic and extrinsic evaluation measures for machine translation and/or summarization},
  pages={65--72},
  year={2005}
}

@article{lin2023neuronpedia,
  title={Neuronpedia: Interactive reference and tooling for analyzing neural networks},
  author={Lin, Johnny and Bloom, Joseph},
  journal={Software available from neuronpedia. org},
  year={2023}
}

@article{conmy2023towards,
  title={Towards automated circuit discovery for mechanistic interpretability},
  author={Conmy, Arthur and Mavor-Parker, Augustine and Lynch, Aengus and Heimersheim, Stefan and Garriga-Alonso, Adri{\`a}},
  journal={Advances in Neural Information Processing Systems},
  volume={36},
  pages={16318--16352},
  year={2023}
}

@article{sanders2012causal,
  title={Causal connectives in discourse: A cross-linguistic perspective},
  author={Sanders, Ted and Stukker, Ninne},
  journal={Journal of pragmatics},
  volume={44},
  number={2},
  pages={131--137},
  year={2012},
  publisher={ELSEVIER SCIENCE BV}
}

@article{adam2014method,
  title={A method for stochastic optimization},
  author={Adam, Kingma DP Ba J and others},
  journal={arXiv preprint arXiv:1412.6980},
  volume={1412},
  number={6},
  year={2014}
}

@misc{nanda2022transformerlens,
    title = {TransformerLens},
    author = {Neel Nanda and Joseph Bloom},
    year = {2022},
    howpublished = {\url{https://github.com/TransformerLensOrg/TransformerLens}},
}

@article{heit2010relations,
  title={Relations between inductive reasoning and deductive reasoning.},
  author={Heit, Evan and Rotello, Caren M},
  journal={Journal of Experimental Psychology: Learning, Memory, and Cognition},
  volume={36},
  number={3},
  pages={805},
  year={2010},
  publisher={American Psychological Association}
}

@misc{openai2025gptoss120bgptoss20bmodel,
      title={gpt-oss-120b \& gpt-oss-20b Model Card}, 
      author={OpenAI},
      year={2025},
      eprint={2508.10925},
      archivePrefix={arXiv},
      primaryClass={cs.CL},
      url={https://arxiv.org/abs/2508.10925}, 
}

@misc{marks2024dictionary_learning,
  title = {Dictionary Learning},
  author = {Samuel Marks and Adam Karvonen and Aaron Mueller},
  year = {2024},
  howpublished = {\url{https://github.com/saprmarks/dictionary_learning}},
  note = {GitHub repository}
}

\appendix

\section{Datasets}
\label{sec:ds}

We extract the reasoning knowledge vectors, validate our linear assumption and refine these vectors based on complementary hypothesis using three logical reasoning datasets: JustLogic \cite{chen2025justlogic}, DEER \cite{yang2024language}, and ART \cite{bhagavatula2019abductive}:

\noindent\textbf{JustLogic} \cite{chen2025justlogic} provides logically structured arguments constructed without relying on external world knowledge, ensuring that models must perform inference rather than recall memorized facts. Each instance consists of a set of formal premises paired with a conclusion that must logically follow. In our experiments, we adopt the official dataset split, which contains 4900 training and 1050 test data points for vector extraction, complementary refinement and evaluation.

\noindent\textbf{DEER} \cite{yang2024language} is a natural-language inductive reasoning dataset in which models are required to infer general rules from concrete textual examples. Each data point contains several observed facts along with a corresponding rule template and the ground truth rule that must be induced. The dataset captures a wide range of natural textual patterns and serves as a realistic benchmark for assessing rule induction. Following the official setup, we use the provided training and test splits (438/762 for train/test) in our experiments.

\noindent\textbf{ART} \cite{bhagavatula2019abductive} is a large-scale abductive reasoning benchmark focused on commonsense explanation generation. Each example presents an observation pair describing a narrative situation, and models must choose the most plausible intermediate hypothesis that links them. The dataset reflects human-like abductive reasoning by emphasizing plausible explanation inference rather than surface-level pattern matching. The dataset consists of 170K training examples and 1.5K test examples. We sampled 4K instances from the training set and used the entire test split from the original benchmark for all experiments.

\section{Implementation Details}
\label{sec:id}

Since we need the SAEs for our complementary refinement and further analysis, we choose Llama3.1-8B-it \cite{grattafiori2024llama} and Gemma-2-9B-it \cite{team2024gemma} as our evaluation LLMs. And since SAEs trained on the middle-layer residual stream of base models generally transfer well to the corresponding instruction fine-tuning models \cite{kissane2024sae}, we choose llama\_scope\_lxr\_8x from Llama Scope \cite{lieberum2024gemma} and gemma-scope-9b-pt-res-canonical from Gemma Scope \cite{he2024llama} to extract features from Llama3.1-8B-it and Gemma-2-9B-it respectively. For GPT-OSS-20B, we use saes-gpt-oss-20b \cite{marks2024dictionary_learning}.

Based on experience of previous work \citep{rimsky2024steering, zhanguncovering}, we extract the activations and steering the model on layer 13 for both Llama-3.1-8B-it and Gemma-2-9B-it. Both the activation extraction and knowledge steering is at the generation stage. For each data point, we collected the activations from every token of generated text. And also we perform knowledge steering by adding it at every token position of the generated text after the end of the initial prompt. The max text generation length is set to be 512 tokens. 

All experiments are conducted on NVIDA B200 GPUs. The batch size is 16. In subspace extraction, we set $\varepsilon = 1e-6$, $\tau = 0.9$ and $K = 3000$. In complementary vectors extraction, using Adam as the optimizer \cite{adam2014method}, the learning rate is set to be $1e-3$, $\lambda_{com}$ is $1e-1$ and $\lambda_{sub}$ is $1e-2$. TransformerLens \cite{nanda2022transformerlens} is used for activation patching.

\section{Activation Patching}
\label{sec:ap}

\begin{figure}[t]
    \centering

    \begin{subfigure}{\linewidth}
        \centering
        \includegraphics[width=\linewidth]{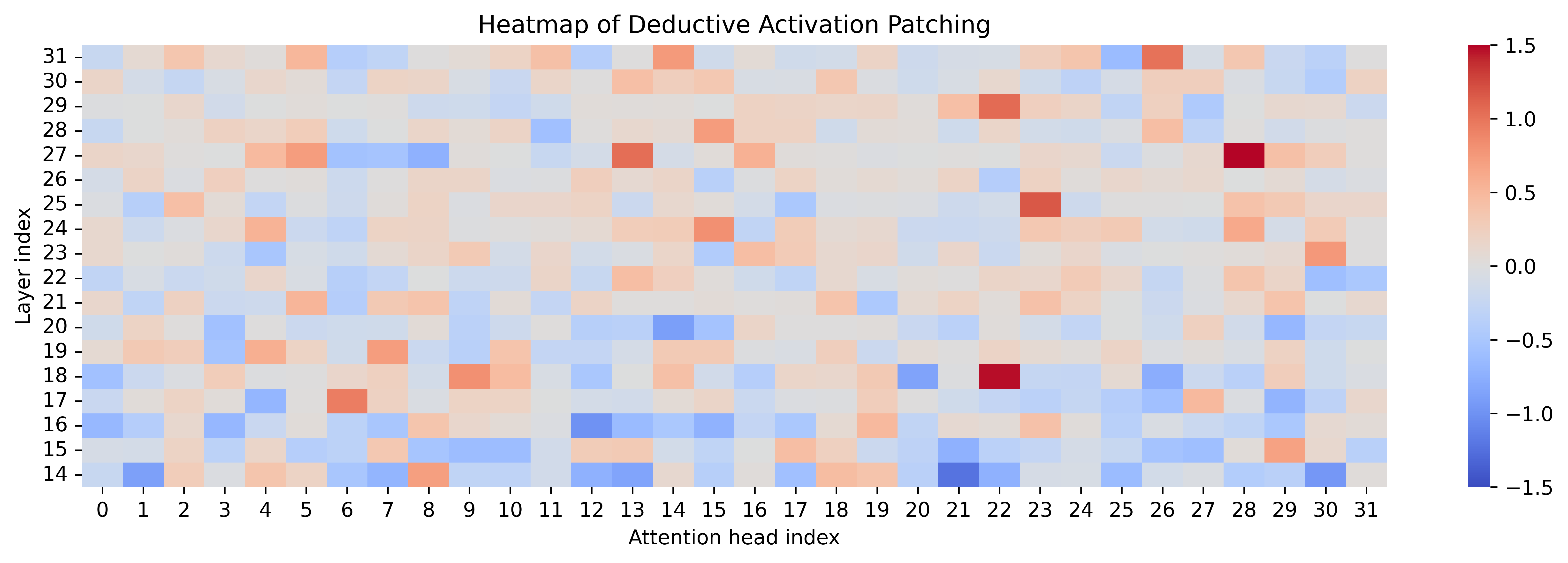}
        \caption*{}
        \label{fig:1a}
    \end{subfigure}

    \vspace{0.6em}

    \begin{subfigure}{\linewidth}
        \centering
        \includegraphics[width=\linewidth]{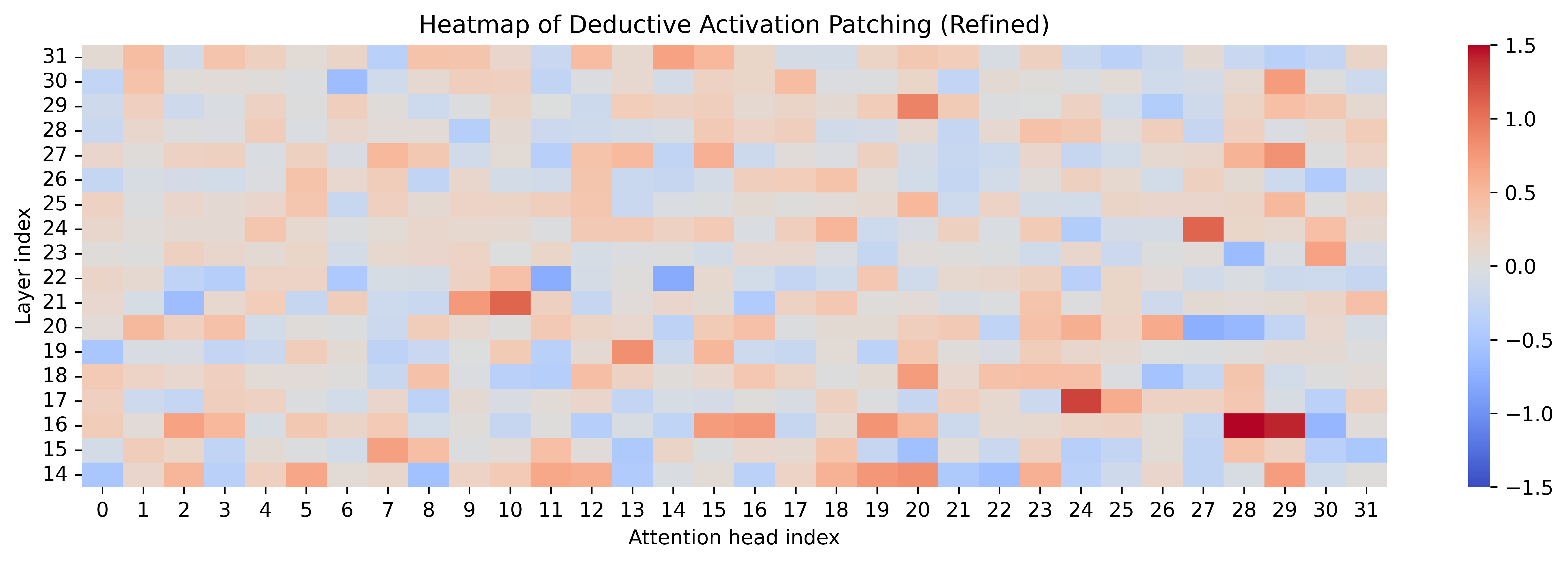}
        \caption*{}
        \label{fig:1d}
    \end{subfigure}

    \vspace{0.6em}

    \begin{subfigure}{\linewidth}
        \centering
        \includegraphics[width=\linewidth]{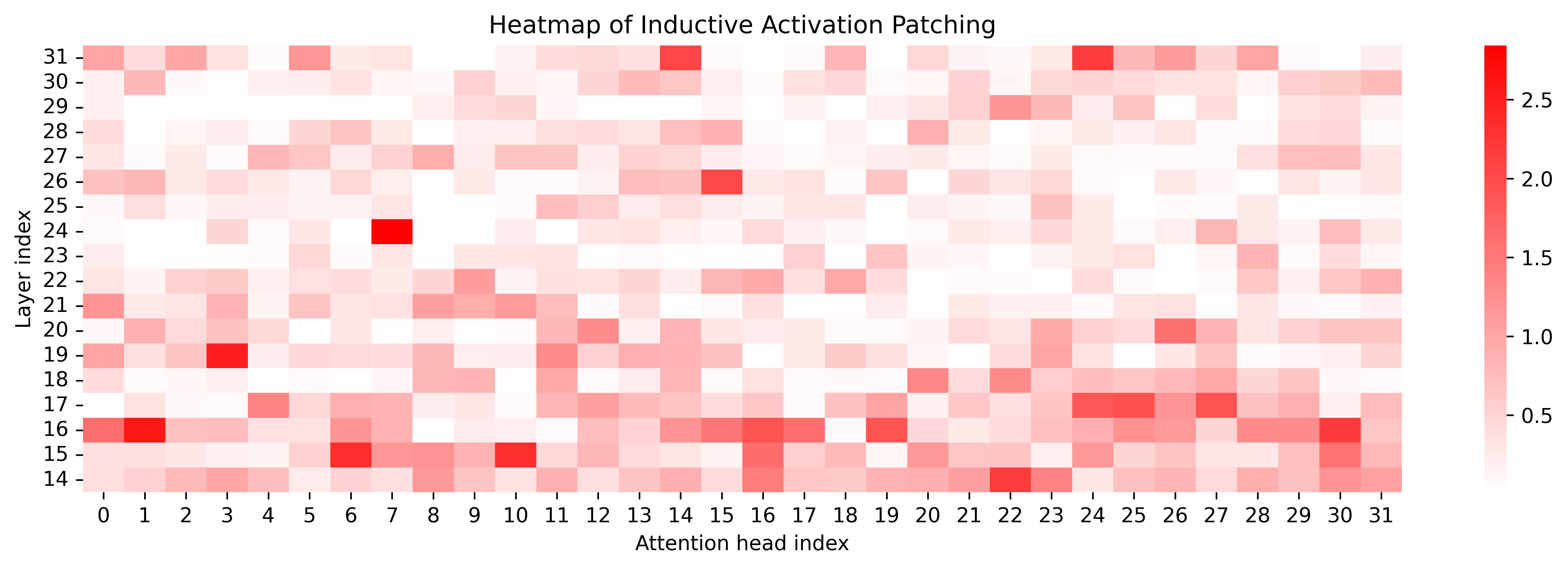}
        \caption*{}
        \label{fig:1b}
    \end{subfigure}

    \vspace{0.6em}

    \begin{subfigure}{\linewidth}
        \centering
        \includegraphics[width=\linewidth]{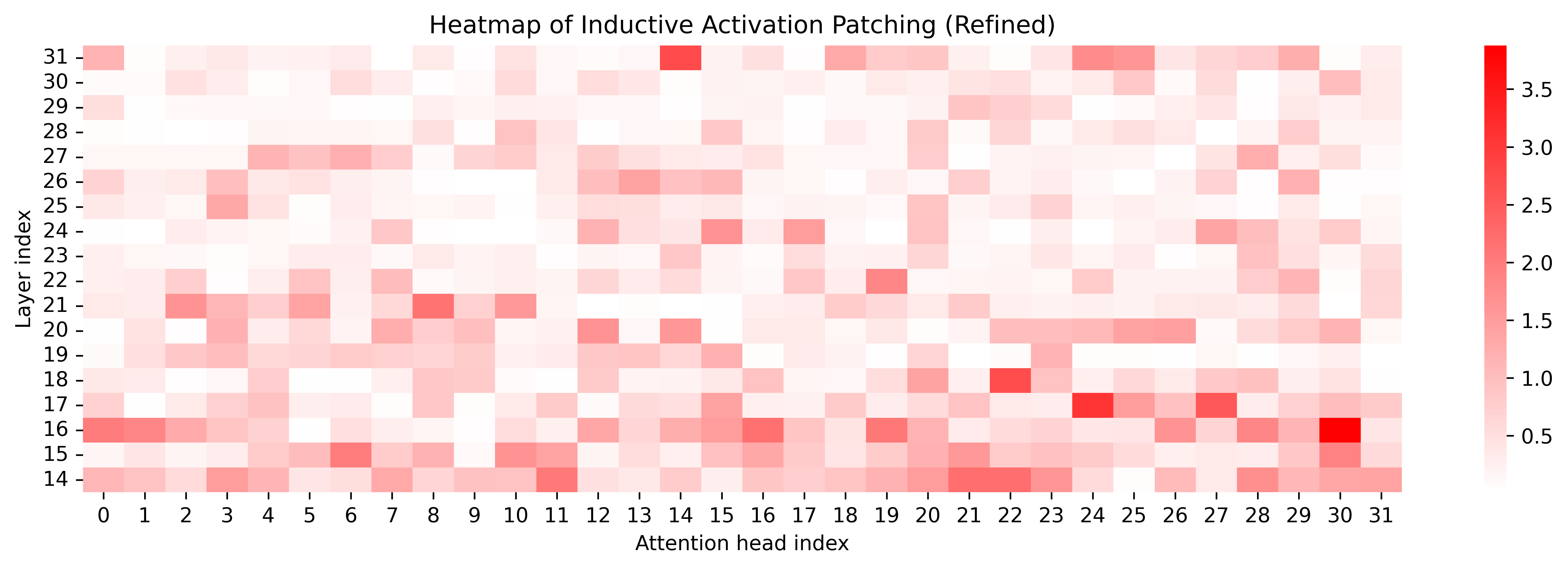}
        \caption*{}
        \label{fig:1e}
    \end{subfigure}

    \vspace{0.6em}

    \begin{subfigure}{\linewidth}
        \centering
        \includegraphics[width=\linewidth]{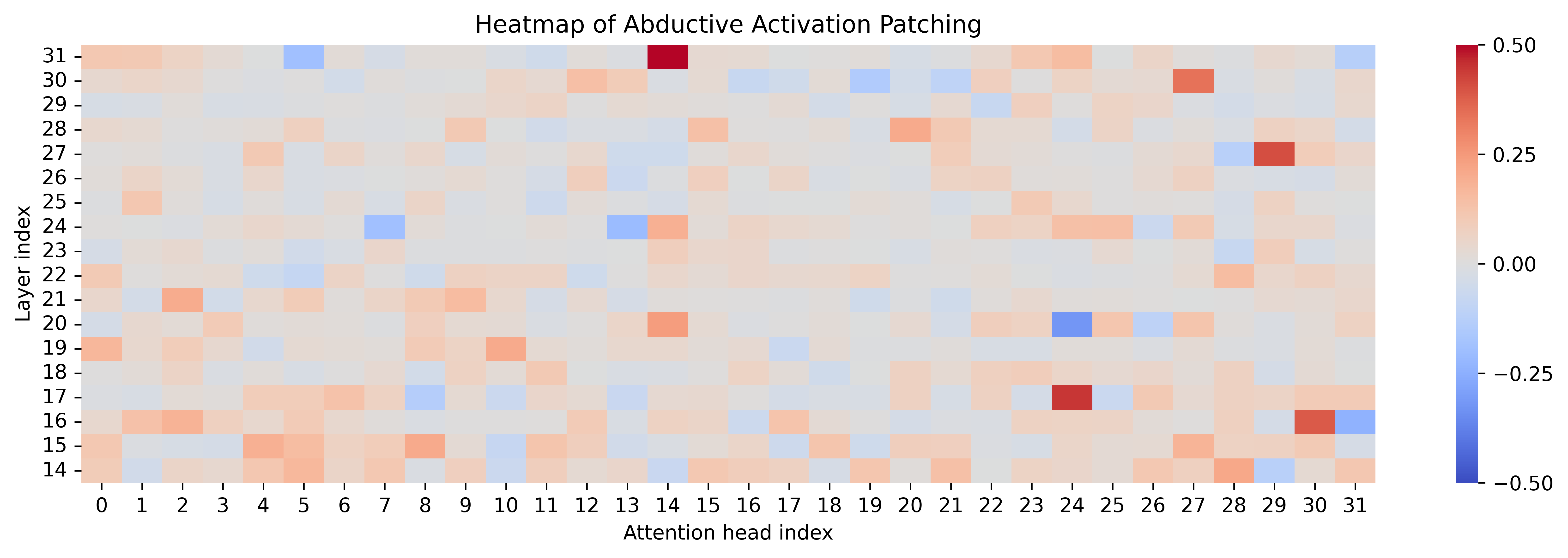}
        \caption*{}
        \label{fig:1c}
    \end{subfigure}

    \vspace{0.6em}

    \begin{subfigure}{\linewidth}
        \centering
        \includegraphics[width=\linewidth]{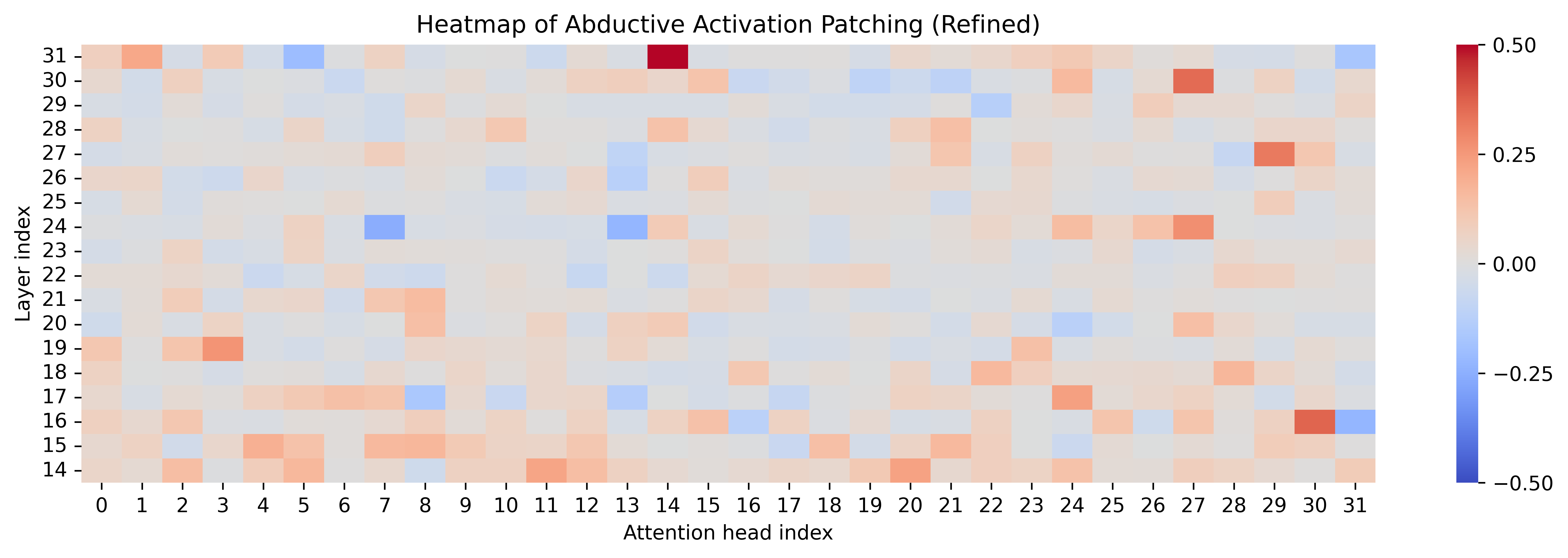}
        \caption*{}
        \label{fig:1f}
    \end{subfigure}

    \caption{Activation patching heatmaps. The top-to-bottom panels show naive steering followed by refined steering for deductive, inductive, and abductive reasoning. Each unit denotes the logit difference of the answer token or the semantic difference in hidden states between patched and unpatched executions.}
    \label{fig:att_patching}
\end{figure}

Since we applied the extracted reasoning knowledge vectors at layer 13, we patch the attention heads from layer 14 to the final layer to observe the effects of our knowledge vectors. First, we run the model without any intervention to cache the orignal attention heads' activations. At the intervening run, we follow the steering style and patch with corresponding cached activations at each generated step. For deductive and abductive reasoning, we choose the answer token logit difference as the metric to evaluate the the effect of knowledge vector on specific attention head. Since there's no ground truth answer token for inductive reasoning, we instead use the last token hidden states semantic difference to evaluate it. The results are shown in Figure~\ref{fig:att_patching}, which illustrating the influence of different reasoning vectors on models' internal circuit.

\section{Prompt Templates}
\label{sec:pt}

Following the prompt formats provided in the original papers of each dataset, we construct positive and negative prompt templates for deductive, inductive, and abductive reasoning. These templates are designed to elicit positive and negative activations corresponding to each reasoning type, enabling us to extract reasoning-aligned activation patterns for subsequent vector construction and analysis.

\subsection{Deductive Reasoning}

\noindent\textbf{Positive.} "You are given a paragraph of premises, followed by a statement. Perform deductive logical reasoning with propositional logic on the paragraph to determine the truth value of the statement.

\noindent Here is the list of argument forms:

\noindent- Modus Ponens

\noindent- Modus Tollens

\noindent- Hypothetical Syllogism

\noindent- Disjunctive Syllogism

\noindent- Reductio ad absurdum

\noindent- Constructive Dilemma

\noindent- Disjunction Elimination

\noindent You must answer with either one of the 3 options:

\noindent- TRUE: When the premises in the paragraph lead to the statement

\noindent- FALSE: When the premises in the paragraph directly contradict the statement

\noindent- UNCERTAIN: When the premises in the paragraph neither support nor contradict the statement

\noindent Your answer should be short and clear, containing:

\noindent 1. Do not repeat paragraph content. Instead, give a brief logical reasoning process directly.

\noindent 2. Final answer: one of "TRUE", "FALSE", or "UNCERTAIN" (on a new line, quoted).

\noindent Use only the information in the paragraph. Assume all premises are true.

\noindent-----

\noindent Paragraph: \{PARAGRAPH\}

\noindent Statement: \{STATEMENT\}

\noindent Answer:"

\noindent\textbf{Negative.} "You are given a paragraph of premises, followed by a statement. Determine the truth value of the statement based on the paragraph.

\noindent You must answer with either one of the 3 options:

\noindent- TRUE: When the premises in the paragraph lead to the statement

\noindent- FALSE: When the premises in the paragraph directly contradict the statement

\noindent- UNCERTAIN: When the premises in the paragraph neither support nor contradict the statement

\noindent Your answer should be short and clear, containing:

\noindent 1. Do not repeat paragraph content.

\noindent 2. Final answer: one of "TRUE", "FALSE", or "UNCERTAIN" (on a new line, quoted).

\noindent Use only the information in the paragraph. Assume all premises are true.

\noindent-----

\noindent Paragraph: \{PARAGRAPH\}

\noindent Statement: \{STATEMENT\}

\noindent Answer:"

\subsection{Inductive Reasoning}

\noindent\textbf{Positive.} "You are a reasoning assistant capable of inductive generalization. Your task is to observe several specific facts and infer a general rule that could have produced them. Focus on finding a pattern that explains all examples and write the rule that satisfies the rule template and the given facts.

\noindent Do not include '\_' in generation. 

\noindent Fact: \{FACT\}

\noindent Rule Template: \{TEMPLATE\}"

\noindent\textbf{Negative.} "Generate a rule in the LAST line that satisfies the rule template and the given facts. 

\noindent Do not include '\_' in generation. 

\noindent Fact: \{FACT\}

\noindent Rule Template: \{TEMPLATE\}"

\subsection{Abductive Reasoning}

\noindent\textbf{Positive.} "You are an expert in abductive logical reasoning. Given the following two observations, your task is to carefully evaluate two hypotheses and determine which one provides a better causal explanation.

\noindent Observations:

\noindent 1. \{OBS1\}

\noindent 2. \{OBS2\}

\noindent Hypotheses:

\noindent A. \{HYP1\}

\noindent B. \{HYP2\}

\noindent For each hypothesis, consider whether it explains both observations plausibly and logically. Think carefully and explain your reasoning briefly.

\noindent Then, answer the following question:

\noindent Which hypothesis (A or B) more plausibly explains the observations?

\noindent Format your answer as:

\noindent Reasoning: <your reasoning here>

\noindent Answer: <A or B>"

\noindent\textbf{Negative.} "You are given two observations that describe a situation, and two hypotheses that attempt to explain what happened.

\noindent Your task is to determine which hypothesis is more plausible given the two observations.

\noindent Observations:

\noindent 1. \{OBS1\}

\noindent 2. \{OBS2\}

\noindent Hypotheses:

\noindent A. \{HYP1\}

\noindent B. \{HYP2\}

\noindent Explain your reasoning briefly.

\noindent Then, answer the following question:

\noindent Which hypothesis (A or B) more plausibly explains the observations?

\noindent Format your answer as:

\noindent Reasoning: <your reasoning here>

\noindent Answer: <A or B>"





\section{Sensitivity Analysis}
\label{sec:ave}

To examine the robustness of the refinement objective to the choice of hyperparameters, we conduct a sensitivity analysis on Llama-3.1-8B-it by varying the two loss weights in Equation~\ref{eq:total_loss}. Specifically, we consider $\lambda_{\text{com}} \in \{10^{-2}, 10^{-1}, 1.0\}$ and $\lambda_{\text{sub}} \in \{10^{-1}, 10^{-2}, 10^{-3}\}$, and summarize the resulting complementary-steering performance in Table~\ref{tab:sensitivity}. Across these settings, performance varies only marginally for deductive, inductive, and abductive reasoning, indicating that the method operates in a relatively stable regime rather than depending on a sharp hyperparameter optimum.

\begin{table}[t]
\centering
\small
\begin{tabular}{ccc}
\toprule
 \textbf{Inductive} & \textbf{Deductive} & \textbf{Abductive} \\
\midrule
27.55$\pm$0.41 & 56.46$\pm$0.85 & 40.95$\pm$0.62 \\
\bottomrule
\end{tabular}

\caption{Summary of sensitivity analysis results on Llama-3.1-8B-it over different choices of $\lambda_{\text{com}}$ and $\lambda_{\text{sub}}$. We report the mean and standard deviation of complementary-steering performance across all tested settings. }
\label{tab:sensitivity}
\end{table}



\end{document}